\pdfoutput=1
\documentclass[11pt,a4paper]{article}
\usepackage[table,x11names,dvipsnames,table]{xcolor}
\usepackage[hyperref]{acl2017}
\usepackage{amsmath}
\usepackage{times}
\usepackage{latexsym}
\usepackage[colorinlistoftodos,prependcaption,textsize=tiny]{todonotes}
\usepackage{enumitem}
\usepackage{array}
\usepackage{multirow}
\usepackage{subfig}
\usepackage{siunitx}
\usepackage{tabu}
\usepackage[compact]{titlesec}
\usepackage[activate={true,nocompatibility},final,tracking=true,kerning=true,factor=1100,stretch=10,shrink=10]{microtype}

\definecolor{lightcornflowerblue}{rgb}{0.6, 0.81, 0.93}
\definecolor{palecornflowerblue}{rgb}{0.67, 0.8, 0.94}
\definecolor{lightskyblue}{rgb}{0.53, 0.81, 0.98}
\definecolor{lightslategray}{rgb}{0.47, 0.53, 0.6}
\definecolor{slategray}{rgb}{0.44, 0.5, 0.56}
\definecolor{dimgray}{rgb}{0.41, 0.41, 0.41}
\definecolor{corn}{rgb}{0.98, 0.93, 0.36}
\definecolor{palespringbud}{rgb}{0.93, 0.92, 0.74}
\definecolor{palesilver}{rgb}{0.79, 0.75, 0.73}
\definecolor{verylightgrey}{rgb}{0.90, 0.90, 0.90}
\definecolor{palegreen}{rgb}{0.6, 0.98, 0.6}

\newcolumntype{H}{>{\setbox0=\hbox\bgroup}c<{\egroup}@{}}

\aclfinalcopy 
\def\aclpaperid{S17-2001} 

\title{SemEval-2017 Task 1: Semantic Textual Similarity \\ Multilingual and Cross-lingual Focused Evaluation}

\author{
Daniel Cer\textsuperscript{$a$}, 
Mona Diab\textsuperscript{$b$},
Eneko Agirre\textsuperscript{$c$}, 
\\\rm\textbf{I\~nigo Lopez-Gazpio\textsuperscript{$c$},
                  and Lucia Specia\textsuperscript{$d$}}
\\\AND
{\rm\textsuperscript{$a$}Google Research}\\Mountain View, CA \And {\rm\textsuperscript{$b$}George Washington University}\\Washington, DC  \AND 
{\rm\textsuperscript{$c$}University of the Basque Country}\\Donostia, Basque Country \And {\rm\textsuperscript{$d$}University of Sheffield}\\Sheffield, UK}

\date{}

\begin{document}
\maketitle
\begin{abstract}
  Semantic Textual Similarity (STS) measures the meaning similarity of sentences. Applications include machine translation (MT), summarization, generation, question answering (QA), short answer grading, semantic search, dialog and conversational systems.
  The STS shared task is a venue for assessing the current state-of-the-art.
  The 2017 task focuses on multilingual and cross-lingual pairs with one sub-track exploring MT quality estimation (MTQE) data. The task obtained strong participation from 31 teams, with 17 participating in \emph{all  language tracks}. We summarize performance and review a selection of well performing methods. Analysis highlights common errors, providing insight into the limitations of existing models. To support ongoing work on semantic representations, the {\em STS Benchmark} is introduced as a new shared training and evaluation set carefully selected from the corpus of English STS shared task data (2012-2017).
\end{abstract}

\section{Introduction}

Semantic Textual Similarity (STS) assesses the degree to which two sentences are semantically equivalent to each other.
The STS task is motivated by the observation that accurately modeling the meaning similarity of sentences is a foundational language understanding problem relevant to numerous applications including: machine translation (MT), summarization, generation, question answering (QA), short answer grading, semantic search, dialog and conversational systems. STS enables the evaluation of techniques from a diverse set of domains against a shared interpretable performance criteria. Semantic inference tasks related to STS include textual entailment \cite{Bentivogli2016, snli:emnlp2015, dagan2009}, semantic relatedness \cite{Bentivogli2016} and paraphrase detection \cite{xu2015semeval, ganitkevitch2013ppdb, :msrpp}. STS differs from both textual entailment and paraphrase detection in that it captures {\it gradations of meaning overlap} rather than making binary classifications of particular relationships. While semantic relatedness expresses a graded semantic relationship as well, it is non-specific about the nature of the relationship with contradictory material still being a candidate for a high score (e.g., ``night" and ``day" are highly related but not particularly similar).

To encourage and support research in this area, the STS shared task has been held annually since 2012, providing a venue for evaluation of state-of-the-art algorithms and models \cite{agirre-EtAl:2012:STARSEM-SEMEVAL,agirre-EtAl:2013:*SEM1,agirre-EtAl:2014:SemEval,agirre-EtAl:2015:SemEval,agirre-EtAl:2016:SemEval1}. During this time, diverse similarity methods and data sets\footnote{i.a., news headlines, video and image descriptions, glosses from lexical resources including WordNet \cite{Miller1995,Fellbaum:98}, FrameNet \cite{Baker:98}, OntoNotes \cite{Hovy:06}, web discussion fora, plagiarism, MT post-editing and Q\&A data sets. Data sets are summarized on: \url{http://ixa2.si.ehu.es/stswiki}.} have been explored.
Early methods focused on lexical semantics, surface form matching and basic syntactic similarity \cite{bar-EtAl:2012:STARSEM-SEMEVAL,vsaric-EtAl:2012:STARSEM-SEMEVAL,jimenez-becerra-gelbukh:2012:STARSEM-SEMEVAL1}. During subsequent evaluations, strong new similarity signals emerged, such as \newcite{Sultan2015}'s alignment based method. More recently, deep learning became competitive with top performing feature engineered systems \cite{He2016}. The best performance tends to be obtained by ensembling feature engineered and deep learning models \cite{Rychalska2016}. 

Significant research effort has focused on STS over English sentence pairs.\footnote{The 2012 and 2013 STS tasks were English only. The 2014 and 2015 task included a Spanish track and 2016 had a pilot track on cross-lingual Spanish-English STS. The English tracks attracted the most participation and have the largest use of the evaluation data in ongoing research.} English STS is a well-studied problem, with state-of-the-art systems often achieving 70 to 80\% correlation with human judgment. To promote progress in other languages, the 2017 task emphasizes performance on Arabic and Spanish as well as cross-lingual pairings of English with material in Arabic, Spanish and Turkish. The \textit{primary} evaluation criteria combines performance on all of the different language conditions except English-Turkish, which was run as a surprise language track. Even with this departure from prior years, the task attracted 31 teams producing 84 submissions.

STS shared task data sets have been used extensively for research on sentence level similarity and semantic representations (i.a., \newcite{sif2017,Conneau2017,Mu2017,pgj2017unsup,Wieting2017,he-lin:2016:N16-1,hill2016,kenter2016,Lau2016,wieting-EtAl:2016:EMNLP2016,Wieting2016,he-gimpel-lin:2015:EMNLP,pham2015}). To encourage the use of a common evaluation set for assessing new methods, we present the STS Benchmark, a publicly available selection of data from English STS shared tasks (2012-2017).

\section{Task Overview}
 
STS is the assessment of pairs of sentences according
to their degree of semantic similarity. The task involves producing real-valued similarity scores for sentence pairs. Performance is measured by the Pearson correlation of machine scores with human judgments. 
 The ordinal scale in Table \ref{fig:annotationcore} guides human annotation, ranging from 0 for no meaning overlap to 5 for meaning equivalence. Intermediate values reflect interpretable levels of partial overlap in meaning. The annotation scale is designed to be accessible by reasonable human judges without any formal expertise in linguistics. Using reasonable human interpretations of natural language semantics was popularized by the related textual entailment task \cite{dagan2009}. The resulting annotations reflect both pragmatic and world knowledge and are more interpretable and useful within downstream systems.
 
\begin{table}[ht!]
\begin{center}
\small
{\renewcommand{\arraystretch}{1.25}
\begin{tabular}{|c|p{0.88\linewidth}|}
\hline
\multirow{4}{*}{5}& \emph{The two sentences are completely equivalent, as they mean the same thing.}\\
\cline{2-2}
& The bird is bathing in the sink. \newline Birdie is washing itself in the water basin. \\
\hline
\multirow{4}{*}{4}  & \emph{The two sentences are mostly equivalent, but some {\it unimportant} details differ.}\\
\cline{2-2}
& Two boys on a couch are playing video games. \newline Two boys are playing a video game. \\
\hline
\multirow{4}{*}{3} & \emph{The two sentences are roughly equivalent, but some {\it important information} differs/missing.}\\
\cline{2-2}
& John said he is considered a witness but not a suspect. \newline ``He is not a suspect anymore.'' John said.  \\
\hline
\multirow{4}{*}{2} & \emph{The two sentences are not equivalent, but share some details.}\\
\cline{2-2}
& They flew out of the nest in groups. \newline They flew into the nest together. \\
\hline
\multirow{4}{*}{1} & \emph{The two sentences are not equivalent, but are on the same topic.}\\
\cline{2-2}
& The woman is playing the violin. \newline The young lady enjoys listening to the guitar. \\
\hline
\multirow{4}{*}{0} & \emph{The two sentences are completely dissimilar.}\\
\cline{2-2}
& The black dog is running through the snow. \newline A race car driver is driving his car through the mud. \\
\hline
\end{tabular}
}
\end{center}
\caption{Similarity scores with explanations and English examples from \newcite{agirre-EtAl:2013:*SEM1}.}
\label{fig:annotationcore}
\end{table}

\section{Evaluation Data}

The Stanford Natural Language Inference (SNLI) corpus \cite{snli:emnlp2015} is the primary evaluation data source with the exception that one of the cross-lingual tracks explores data from the WMT 2014 quality estimation task \cite{bojar-etal_WMT:2014}.\footnote{Previous years of the STS shared task include more data sources. This year the task draws from two data sources and includes a diverse set of languages and language-pairs.} 

Sentences pairs in SNLI derive from Flickr30k image captions \cite{young2014} and are labeled with the entailment relations: entailment, neutral, and contradiction. Drawing from SNLI allows STS models to be evaluated on the type of data used to assess textual entailment methods. However, since entailment strongly cues for semantic relatedness \cite{MARELLI14.363}, we construct our own sentence pairings to deter gold entailment labels from informing evaluation set STS scores. 

Track 4b investigates the relationship between STS and MT quality estimation by providing STS labels for WMT quality estimation data. The data includes Spanish translations of English sentences from a variety of methods including RBMT, SMT, hybrid-MT and human translation. Translations are annotated with the time required for human correction by post-editing and Human-targeted Translation Error Rate (HTER) \cite{snover2006}.\footnote{HTER is the minimal number of edits required for correction of a translation divided by its length after correction.} Participants are not allowed to use the gold quality estimation annotations to inform STS scores.

\begin{table}[t]
\small
\begin{center}
\begin{tabular}{|l|l|r|l|} 
\hline
     Track & Language(s) & Pairs &  Source\\
\hline
  1 & Arabic (ar-ar) & 250 & SNLI \\
  2 & Arabic-English (ar-en) & 250 & SNLI \\
  3 & Spanish (es-es) & 250 & SNLI \\
  4a & Spanish-English (es-en) & 250 & SNLI \\
  4b & Spanish-English (es-en) & 250 & WMT QE \\
  5  & English (en-en) & 250 & SNLI \\
  6 &  Turkish-English (tr-en) &  250 & SNLI \\
\hline
    & Total & 1750 & \\
\hline
\end{tabular}
\end{center}
\caption{STS 2017 evaluation data.
}
\label{tab:sts2017eval-summary}
\end{table}

\subsection{Tracks}
Table \ref{tab:sts2017eval-summary} summarizes the evaluation data by track. The six tracks span four languages: Arabic, English, Spanish and Turkish. 
Track 4 has subtracks with 4a drawing from SNLI and 4b pulling from WMT's quality estimation task. Track 6 is a surprise language track with no annotated training data and the identity of the language pair first announced when the evaluation data was released.

\subsection{Data Preparation}\label{sec:data-collection}

This section describes the preparation of the evaluation data. For SNLI data, this includes the selection of sentence pairs, annotation of pairs with STS labels and the translation of the original English sentences. WMT quality estimation data is directly annotated with STS labels.

\subsection{Arabic, Spanish and Turkish Translation}\label{sec:data-translation}

 Sentences from SNLI are human translated into Arabic, Spanish and Turkish. Sentences are translated independently from their pairs. Arabic translation is provided by CMU-Qatar by native Arabic speakers with strong English skills. Translators are given an English sentence and its Arabic machine translation\footnote{Produced by the Google Translate API.} where they perform post-editing to correct errors. Spanish translation is completed by a University of Sheffield graduate student who is a native Spanish speaker and fluent in English. Turkish translations are obtained from SDL.\footnote{\url{http://www.sdl.com/languagecloud/managed-translation/}} 
 
\subsection{Embedding Space Pair Selection} \label{embeddingspacepairselection}

We construct our own pairings of the SNLI sentences to deter gold entailment labels being used to inform STS scores. The {\it word embedding similarity} selection heuristic from STS 2016 \cite{agirre-EtAl:2016:SemEval1} is used to find interesting pairs. Sentence embeddings are computed as the sum of individual word embeddings, $\mathbf{v}(s) = \sum_{w \in s} \mathbf{v} (w)$.\footnote{We use 50-dimensional GloVe word embeddings \cite{pennington2014glove} trained on a combination of Gigaword 5 \cite{parker-EtAl:2011} and English Wikipedia available at \url{http://nlp.stanford.edu/projects/glove/}.} Sentences with likely meaning overlap are identified using cosine similarity, Eq. (\ref{deepsim}).

\begin{equation}
\label{deepsim}
\text{sim}_{\text{v}}(s_1, s_2) = \frac {\mathbf{v}(s_1) \mathbf{v}(s_2)}{\lVert\mathbf{v}(s_1)\rVert_2 \lVert\mathbf{v}(s_2)\rVert_2}
\end{equation}

\section{Annotation}

Annotation of pairs with STS labels is performed using Crowdsourcing, with the exception of Track 4b that uses a single expert annotator.

\subsection{Crowdsourced Annotations}

Crowdsourced annotation is performed on Amazon Mechanical Turk.\footnote{\url{https://www.mturk.com/}} Annotators examine the STS pairings of English SNLI sentences. STS labels are then transferred to the translated pairs for cross-lingual and non-English tracks.
The annotation instructions and template are identical to \newcite{agirre-EtAl:2016:SemEval1}. 
Labels are collected in batches of 20 pairs with annotators paid \$1 USD per batch. Five annotations are collected per pair. The MTurk {\em master}\footnote{A designation that statistically identifies workers who perform high quality work across a diverse set of tasks.}  qualification is required to perform the task. Gold scores average the five individual annotations.

\subsection{Expert Annotation}

Spanish-English WMT quality estimation pairs for Track 4b are annotated for STS by a University of Sheffield graduate student who is a native speaker of Spanish and fluent in English. This track differs significantly in label distribution and the complexity of the annotation task. Sentences in a pair are translations of each other and tend to be more semantically similar. Interpreting the potentially subtle meaning differences introduced by MT errors is challenging. To accurately assess STS performance on MT quality estimation data, no attempt is made to balance the data by similarity scores.

\begin{table}[t]
\small
\begin{center}
\begin{tabular}{|l|l|r|l|} 
\hline
     Year & Data set         & Pairs &  Source\\
\hline
2012 & MSRpar    & 1500  &  newswire\\
2012 &MSRvid    &     1500 &videos\\
2012 &OnWN      &     750  &glosses\\
2012 &SMTnews   &     750  &WMT eval.\\
2012 &SMTeuroparl &   750  &WMT eval.\\
\hline
2013 &HDL &     750  &newswire\\
2013 &FNWN      & 189      & glosses\\
2013 &OnWN      & 561     & glosses\\
2013 &SMT       &750      & MT eval.\\
\hline
2014 &HDL &     750  &newswire headlines\\
2014 &OnWN      & 750     & glosses\\
2014 &Deft-forum       &450      &forum posts\\
2014 &Deft-news     &300    &news summary\\
2014 &Images       &750      & image descriptions\\
2014 &Tweet-news &750      & tweet-news pairs\\
\hline
2015 &HDL &     750  &newswire headlines\\
2015 &Images       &750      & image descriptions\\
2015 &Ans.-student &750      & student answers\\
2015 &Ans.-forum &375      & Q\&A forum answers\\
2015 &Belief &375      & committed belief\\
\hline
2016 & HDL & 249 & newswire headlines\\
2016 & Plagiarism & 230 & short-answer plag. \\
2016 & post-editing & 244 & MT postedits \\
2016 & Ans.-Ans. & 254 & Q\&A forum answers\\
2016 & Quest.-Quest. & 209 & Q\&A forum questions\\
\hline
2017 & Trial & 23 & Mixed STS 2016 \\
\hline
\end{tabular}
\end{center}
\caption{English training data.
}
\label{tab:en-summary}
\end{table}

\begin{table}[t]
\small
\begin{center}
\begin{tabular}{|l|l|r|l|} 
\hline
     Year & Data set         & Pairs &  Source\\
\hline
2014 & Trial & 56 & \\
2014 & Wiki  & 324 & Spanish Wikipedia \\
2014 & News  & 480 & Newswire \\
2015 & Wiki  & 251 & Spanish Wikipedia \\
2015 & News  & 500 & Newswire \\
2017 & Trial & 23 & Mixed STS 2016 \\
\hline
\end{tabular}
\end{center}
\caption{Spanish training data.
}
\label{tab:es-summary}
\end{table}

\begin{table}[t]
\small
\begin{center}
\begin{tabular}{|l|l|r|l|} 
\hline
     Year & Data set         & Pairs &  Source\\
\hline
2016 & Trial & 103 & Sampled $\leq$ 2015 STS\\
2016 & News    & 301  & en-es news articles \\
2016 & Multi-source   & 294 & en news headlines, \\
     &                & &  short-answer plag., \\
     &                & & MT postedits,   \\
     &               & & Q\&A forum answers, \\
     &               & & Q\&A forum questions \\
2017 & Trial & 23 & Mixed STS 2016 \\
2017 & MT & 1000 & WMT13 Translation Task \\
\hline
\end{tabular}
\end{center}
\caption{Spanish-English training data.
}
\label{tab:es-en-summary}
\end{table}

\begin{table}[t]
\small
\begin{center}
\begin{tabular}{|l|l|r|l|} 
\hline
     Year & Data set         & Pairs &  Source\\
\hline
2017 & Trial & 23 & Mixed STS 2016 \\
2017 & MSRpar & 510 & newswire\\
2017 & MSRvid & 368 & videos\\
2017 & SMTeuroparl & 203 & WMT eval.\\
\hline
\end{tabular}
\end{center}
\caption{Arabic training data.
}
\label{tab:ar-summary}
\end{table}

\begin{table}[t]
\small
\begin{center}
\begin{tabular}{|l|l|r|l|} 
\hline
     Year & Data set         & Pairs &  Source\\
\hline
2017 & Trial & 23 & Mixed STS 2016 \\
2017 & MSRpar & 1020 & newswire\\
2017 & MSRvid & 736 & videos\\
2017 & SMTeuroparl & 406 & WMT eval.\\
\hline
\end{tabular}
\end{center}
\caption{Arabic-English training data.
}
\label{tab:ar-en-summary}
\end{table}

\begin{table}[t]
\small
\begin{center}
\begin{tabular}{|l|l|r|l|} 
\hline
     Year & Data set         & Pairs &  Source\\
\hline
2017 & MSRpar & 1039 & newswire\\
2017 & MSRvid & 749 & videos\\
2017 & SMTeuroparl & 422 & WMT eval.\\
\hline
\end{tabular}
\end{center}
\caption{Arabic-English parallel data.
}
\label{tab:ar-en-parallel}
\end{table}

\begin{table}[htp]
\begin{center}
\centering
\begin{tabular}{|c|l|c|}
\hline
  Track & Language(s) & Participants \\ 
\hline
  1 & Arabic & 49 \\
  2 & Arabic-English & 45 \\
  3 & Spanish & 48 \\
  4a & Spanish-English & 53 \\
  4b & Spanish-English MT & 53 \\
  5  & English & 77 \\
  6  & Turkish-English & 48 \\
\hline
  Primary & All except Turkish & 44 \\
\hline\end{tabular}

\end{center}
\caption{Participation by shared task track.}
\label{tab:participation}
\end{table}

\section{Training Data}
The following summarizes the training data: Table \ref{tab:en-summary} English; Table \ref{tab:es-summary} Spanish;\footnote{Spanish data from 2015 and 2014 uses a 5 point scale that collapses STS labels 4 and 3, removing the distinction between unimportant and important details.} Table \ref{tab:es-en-summary} Spanish-English; Table \ref{tab:ar-summary} Arabic; and Table \ref{tab:ar-en-summary} Arabic-English. Arabic-English parallel data is supplied by translating English training data, Table \ref{tab:ar-en-parallel}. 

English, Spanish and Spanish-English training data pulls from prior STS evaluations. Arabic and Arabic-English training data is produced by translating a subset of the English training data and transferring the similarity scores. For the MT quality estimation data in track 4b, Spanish sentences are translations of their English counterparts, differing substantially from existing Spanish-English STS data. We release one thousand new Spanish-English STS pairs sourced from the 2013 WMT translation task and produced by a phrase-based Moses SMT system \cite{bojar-EtAl:2013:WMT}. The data is expert annotated and has a similar label distribution to the track 4b test data with 17\% of the pairs scoring an STS score of less than 3, 23\% scoring 3, 7\% achieving a score of 4 and 53\% scoring 5. 

\subsection{Training vs. Evaluation Data Analysis}

Evaluation data from SNLI tend to have sentences that are slightly shorter than those from prior years of the STS shared task, while the track 4b MT quality estimation data has sentences that are much longer. The track 5 English data has an average sentence length of 8.7 words, while the English sentences from track 4b have an average length of 19.4. The English training data has the following average lengths: 2012 10.8 words; 2013 8.8 words (excludes restricted SMT data); 2014 9.1 words; 2015 11.5 words; 2016 13.8 words.

Similarity scores for our pairings of the SNLI sentences are slightly lower than recent shared task years and much lower than early years. The change is attributed to differences in data selection and filtering. The average 2017 similarity score is 2.2 overall and 2.3 on the track 7 English data. Prior English data has the following average similarity scores: 2016 2.4; 2015 2.4; 2014 2.8; 2013 3.0; 2012 3.5. Translation quality estimation data from track 4b has an average similarity score of 4.0. 

\section{System Evaluation}

This section reports participant evaluation results for the SemEval-2017 STS shared task.

\subsection{Participation}

The task saw strong participation with 31 teams producing 84 submissions. 17 teams provided 44 systems that participated in all tracks. Table \ref{tab:participation} summarizes participation by track. Traces of the focus on English are seen in 12 teams participating just in track 5, English. Two teams participated exclusively in tracks 4a and 4b, Spanish-English. One team took part solely in track 1, Arabic.

\subsection{Evaluation Metric}

Systems are evaluated on each track by their Pearson correlation with gold labels. The overall ranking averages the correlations across tracks 1-5 with tracks 4a and 4b individually contributing.

\subsection{CodaLab}

As directed by the SemEval workshop organizers, the CodaLab research platform hosts the task.\footnote{\url{https://competitions.codalab.org/competitions/16051}}

\subsection{Baseline}
The baseline is the cosine of binary sentence vectors with each dimension representing whether an individual word appears in a sentence.\footnote{Words obtained using Arabic (ar), Spanish (es) and English (en) Treebank tokenizers.} For cross-lingual pairs, non-English sentences are translated into English using state-of-the-art machine translation.\footnote{\url{http://translate.google.com}} The baseline achieves an average correlation of 53.7 with human judgment on tracks 1-5 and would rank 23\textsuperscript{rd} overall out the 44 system submissions that participated in all tracks.

\begin{table*}[htp!!]
\fontsize{6.5pt}{7pt}
\selectfont
\begin{center}
\centering
\rowcolors{5}{}{palecornflowerblue}
\taburulecolor{dimgray}
\definecolor{goldenpoppy}{rgb}{0.99, 0.76, 0.0}
	
\begin{tabu}{|HHl| *{8}{S[table-format=2.2]|}}
\hline
\rowcolor{Wheat1}
&  & \multicolumn{1}{c}{} & \multicolumn{1}{|c|}{} & \multicolumn{1}{|c|}{Track 1} & \multicolumn{1}{|c|}{Track 2} & \multicolumn{1}{|c|}{Track 3} & \multicolumn{1}{|c|}{Track 4a} & \multicolumn{1}{|c|}{Track 4b} & \multicolumn{1}{|c|}{Track 5} & \multicolumn{1}{|c|}{Track 6}\\
\rowcolor{Wheat1}
run ID & username & \multicolumn{1}{c|}{Team} & \multicolumn{1}{|c|}{Primary} & \multicolumn{1}{|c|}{AR-AR} & \multicolumn{1}{|c|}{AR-EN} & \multicolumn{1}{|c|}{SP-SP} & \multicolumn{1}{|c|}{SP-EN} & \multicolumn{1}{|c|}{SP-EN-WMT} & \multicolumn{1}{|c|}{EN-EN} & \multicolumn{1}{|c|}{EN-TR}\\ 
\hline
386611 & lanman & ECNU \cite{tian-EtAl:2017:SemEval} & 73.16 & {\bf 74.40}\hspace{0.55em} & {\bf 74.93} $\bullet$ & {\bf 85.59} $\bullet$ & {\bf 81.31}\hspace{0.55em} & {\bf 33.63}\hspace{0.55em} & {\bf 85.18}\hspace{0.55em} & {\bf 77.06} $\bullet$\\

386610 & lanman & ECNU \cite{tian-EtAl:2017:SemEval} & 70.44 & {\bf 73.80}\hspace{0.55em} & 71.26 & {\bf 84.56}\hspace{0.55em} & 74.95 & {\bf 33.11}\hspace{0.55em} & {\bf 81.81}\hspace{0.55em} & 73.62\\

386608 & lanman & ECNU \cite{tian-EtAl:2017:SemEval} & 69.40 & {\bf 72.71}\hspace{0.55em} & 69.75 & 82.47 & 76.49 & {\bf 26.33}\hspace{0.55em} & {\bf 83.87}\hspace{0.55em} & 74.20\\

387941 & haowu (*) & BIT \cite{wu-EtAl:2017:SemEval1}* & 67.89 & {\bf 74.17}\hspace{0.55em} & 69.65 & {\bf 84.99}\hspace{0.55em} & 78.28 & 11.07 & {\bf 84.00}\hspace{0.55em} & {\bf 73.05}\hspace{0.55em}\\ 

387940 & haowu (*) & BIT \cite{wu-EtAl:2017:SemEval1}* & 67.03 & 75.35 & 70.07 & 83.23 & 78.13 & 7.58 & 81.61 & 73.27\\ 

386849 & haowu & BIT \cite{wu-EtAl:2017:SemEval1} & 66.62 & {\bf 75.43} $\bullet$ & 69.53 & {\bf 82.89}\hspace{0.55em} & 77.61 & 5.84 & {\bf 82.22}\hspace{0.55em} & {\bf 72.80}\hspace{0.55em}\\ 

386540 & Kamikaze & HCTI \cite{shao:2017:SemEval} & 65.98 & {\bf 71.30}\hspace{0.55em} & 68.36 & {\bf 82.63}\hspace{0.55em} & 76.21 & 14.83 & 81.13 & 67.41\\ 

386854 & guidoz & MITRE \cite{henderson-EtAl:2017:SemEval} & 65.90 & {\bf 72.94}\hspace{0.55em} & 67.53 & 82.02 & 78.02 & 15.98 & 80.53 & 64.30\\ 

386852 & guidoz & MITRE \cite{henderson-EtAl:2017:SemEval} & 65.87 & {\bf 73.04}\hspace{0.55em} & 67.40 & 82.01 & 77.99 & 15.74 & 80.48 & 64.41\\ 

386002 & B.Hassan & FCICU \cite{hassan-EtAl:2017:SemEval} & 61.90 & {\bf 71.58}\hspace{0.55em} & 67.82 & {\bf 84.84}\hspace{0.55em} & 69.26 & 2.54 & {\bf 82.72}\hspace{0.55em} & 54.52\\ 

386806 & wlzhuang & neobility \cite{zhuang-chang:2017:SemEval} & 61.71 & 68.21 & 64.59 & 79.28 & 71.69 & 2.00 & 79.27 & 66.96\\ 

386003 & B.Hassan & FCICU \cite{hassan-EtAl:2017:SemEval} & 61.66 & {\bf 71.58}\hspace{0.55em} & 67.81 & {\bf 84.89}\hspace{0.55em} & 68.54 & 2.14 & {\bf 82.80}\hspace{0.55em} & 53.90\\ 

386752 & rekaby & STS-UHH \cite{kohail-salama-biemann:2017:SemEval} & 60.58 & 67.81 & 63.07 & 77.13 & 72.01 & 4.81 & 79.89 & 59.37\\ 

386421 & ruthva & RTV & 60.50 & 67.13 & 55.95 & 74.85 & 70.50 & 7.61 & {\bf 85.41}\hspace{0.55em} & 62.04\\ 

386626 & Kamikaze & HCTI \cite{shao:2017:SemEval} & 59.88 & 43.73 & 68.36 & 67.09 & 76.21 & 14.83 & {\bf 81.56}\hspace{0.55em} & 67.41\\ 

386418 & ruthva & RTV & 59.80 & 66.89 & 54.82 & 74.24 & 69.99 & 7.34 & {\bf 85.41}\hspace{0.55em} & 59.89\\ 

386734 & vijaypal & MatrusriIndia & 59.60 & 68.60 & 54.64 & 76.14 & 71.18 & 5.72 & 77.44 & 63.49\\ 

387475 & rekaby & STS-UHH \cite{kohail-salama-biemann:2017:SemEval} & 57.25 & 61.04 & 59.10 & 72.04 & 63.38 & 12.05 & 73.39 & 59.72\\ 

386803 & StefyD & SEF@UHH \cite{duma-menzel:2017:SemEval} & 56.76 & 57.90 & 53.84 & 74.23 & 58.66 & 18.02 & 72.56 & 62.11\\ 

386435 & StefyD & SEF@UHH \cite{duma-menzel:2017:SemEval} & 56.44 & 55.88 & 47.89 & 74.56 & 57.39 & {\bf 30.69}\hspace{0.55em} & 78.80 & 49.90\\ 

386423 & ruthva & RTV & 56.33 & 61.43 & 48.32 & 68.63 & 61.40 & 8.29 & {\bf 85.47} $\bullet$ & 60.79\\ 

386685 & StefyD & SEF@UHH \cite{duma-menzel:2017:SemEval} & 55.28 & 57.74 & 48.13 & 69.79 & 56.60 & {\bf 34.07} $\bullet$ & 71.86 & 48.78\\ 

386318 & wlzhuang & neobility \cite{zhuang-chang:2017:SemEval} & 51.95 & 13.69 & 62.59 & 77.92 & 69.30 & 0.44 & 75.56 & 64.18\\ 

386559 & wlzhuang & neobility \cite{zhuang-chang:2017:SemEval} & 50.25 & 3.69 & 62.07 & 76.90 & 69.47 & 1.47 & 75.35 & 62.79\\ 

386730 & vijaypal & MatrusriIndia & 49.75 & 57.03 & 43.40 & 67.86 & 55.63 & 8.57 & 65.79 & 49.94\\ 

386791 & thomasp & NLPProxem & 49.02 & 51.93 & 53.13 & 66.42 & 51.44 & 9.96 & 62.56 & 47.67\\ 

386800 & jdbarrow & UMDeep \cite{barrow-peskov:2017:SemEval} & 47.92 & 47.53 & 49.39 & 51.65 & 56.15 & 16.09 & 61.74 & 52.93\\ 

386796 & jeanmarcmarty & NLPProxem & 47.90 & 55.06 & 43.69 & 63.81 & 50.79 & 14.14 & 64.63 & 43.20\\ 

386826 & jdbarrow & UMDeep \cite{barrow-peskov:2017:SemEval} & 47.73 & 45.87 & 51.99 & 51.48 & 52.32 & 13.00 & 62.22 & 57.25\\ 

388044 & cristinae (*) & Lump \cite{espanabonet-barroncedeno:2017:SemEval}* & 47.25 & 60.52 & 18.29 & 75.74 & 43.27 & 1.16 & 73.76 & 58.00\\ 

388046 & cristinae (*) & Lump \cite{espanabonet-barroncedeno:2017:SemEval}* & 47.04 & 55.08 & 13.57 & 76.76 & 48.25 & 11.12 & 72.69 & 51.79\\ 

388045 & cristinae (*) & Lump \cite{espanabonet-barroncedeno:2017:SemEval}* & 44.38 & 62.87 & 18.05 & 73.80 & 44.47 & 1.51 & 73.47 & 36.52\\ 

386831 & jeanmarcmarty & NLPProxem & 40.70 & 53.27 & 47.73 & 0.16 & 55.06 & 14.40 & 66.81 & 47.46\\ 

387486 & Bicici (*) & RTM \cite{biccici:2017:SemEval}* & 36.69 & 33.65 & 17.11 & 69.90 & 60.04 & 14.55 & 54.68 & 6.87\\ 

386787 & dpeskov & UMDeep \cite{barrow-peskov:2017:SemEval} & 35.21 & 39.05 & 37.13 & 45.88 & 34.82 & 5.86 & 47.27 & 36.44\\ 

387487 & Bicici (*) & RTM \cite{biccici:2017:SemEval}* & 32.91 & 33.65 & 0.25 & 56.82 & 50.54 & 13.68 & 64.05 & 11.36\\ 

387485 & Bicici (*) & RTM \cite{biccici:2017:SemEval}* & 32.78 & 41.56 & 13.32 & 48.41 & 45.83 & 23.47 & 56.32 & 0.55\\ 

387457 & bjerva & ResSim \cite{bjerva-ostling:2017:SemEval} & 31.48 & 28.92 & 10.45 & 66.13 & 23.89 & 3.05 & 69.06 & 18.84\\ 

387446 & bjerva & ResSim \cite{bjerva-ostling:2017:SemEval} & 29.38 & 31.20 & 12.88 & 69.20 & 10.02 & 1.62 & 68.77 & 11.95\\ 

387456 & bjerva & ResSim \cite{bjerva-ostling:2017:SemEval} & 21.45 & 0.33 & 10.98 & 54.65 & 22.62 & 1.99 & 50.57 & 9.02\\ 

387465 & ivanvladimir & LIPN-IIMAS \cite{arroyofernandez-mezaruiz:2017:SemEval} & 10.67 & 4.71 & 7.69 & 15.27 & 17.19 & 14.46 & 7.38 & 8.00\\ 

387466 & ivanvladimir & LIPN-IIMAS \cite{arroyofernandez-mezaruiz:2017:SemEval} & 9.26 & 2.14 & 12.92 & 4.58 & 1.20 & 1.91 & 20.38 & 21.68\\ 

386092 & hjpwhu & hjpwhu & 4.80 & 4.12 & 6.39 & 6.17 & 2.04 & 6.24 & 1.14 & 7.53\\ 

387496 & hjpwhu & hjpwhu & 2.94 & 4.77 & 2.04 & 7.63 & 0.46 & 2.57 & 0.69 & 2.46\\

386777 & ferreroj & compiLIG \cite{ferrero-EtAl:2017:SemEval}  &  &  &  &  & {\bf 83.02} $\bullet$ & 15.50 &  & \\ 

386775 & ferreroj & compiLIG \cite{ferrero-EtAl:2017:SemEval} &  &  &  &  & 76.84 & 14.64 &  & \\ 

387239 & ferreroj & compiLIG \cite{ferrero-EtAl:2017:SemEval} &  &  &  &  & 79.10 & 14.94 &  & \\

385817 & nmharjan & DT\_TEAM \cite{maharjan-EtAl:2017:SemEval} &  &  &  &  &  &  & {\bf 85.36}\hspace{0.55em} & \\ 

385818 & nmharjan & DT\_TEAM  \cite{maharjan-EtAl:2017:SemEval} &  &  &  &  &  &  & {\bf 83.60}\hspace{0.55em} & \\ 

385819 & nmharjan & DT\_TEAM  \cite{maharjan-EtAl:2017:SemEval} &  &  &  &  &  &  & {\bf 83.29}\hspace{0.55em} & \\ 

386004 & B.Hassan & FCICU \cite{hassan-EtAl:2017:SemEval} &  &  &  &  &  &  & {\bf 82.17}\hspace{0.55em} & \\ 

386579 & HITInsun & ITNLP–AiKF \cite{liu-EtAl:2017:SemEval1} &  &  &  &  &  &  & {\bf 82.31}\hspace{0.55em} & \\ 

386590 & HITInsun & ITNLP–AiKF \cite{liu-EtAl:2017:SemEval1} &  &  &  &  &  &  & {\bf 82.31}\hspace{0.55em} & \\ 

386569 & HITInsun & ITNLP–AiKF \cite{liu-EtAl:2017:SemEval1} &  &  &  &  &  &  & {\bf 81.59}\hspace{0.55em} & \\  

387244 & pfialho (*) & L2F/INESC-ID \cite{fialho-EtAl:2017:SemEval}* &  &  &  & 76.16 & 1.91 & 5.44 & 78.11 & 2.93\\ 

387243 & pfialho & L2F/INESC-ID \cite{fialho-EtAl:2017:SemEval} &  &  &  &  &  &  & 69.52 & \\ 

386778 & Pfialho (*) & L2F/INESC-ID \cite{fialho-EtAl:2017:SemEval}* &  &  &  & 63.85 & 15.61 & 5.24 & 66.61 & 3.56\\ 

386748 & LIM-LIG & LIM-LIG \cite{nagoudi-ferrero-schwab:2017:SemEval} &  & {\bf 74.63}\hspace{0.55em} &  &  &  &  &  & \\ 

386747 & LIM-LIG & LIM-LIG \cite{nagoudi-ferrero-schwab:2017:SemEval} &  & {\bf 73.09}\hspace{0.55em} &  &  &  &  &  & \\ 

386746 & LIM-LIG & LIM-LIG \cite{nagoudi-ferrero-schwab:2017:SemEval} &  & 59.57 &  &  &  &  &  & \\

386723 & vijaypal & MatrusriIndia &  & 68.60 &  & 76.14 & 71.18 & 5.72 & 77.44 & 63.49\\ 

387920 & cyrilgoutte (*) & NRC* &  &  &  &  & 42.25 & 0.23 &  & \\ 

386804 & cyrilgoutte & NRC &  &  &  &  & 28.08 & 11.33 &  & \\ 

386074 & OkadaNaoya & OkadaNaoya &  &  &  &  &  &  & 77.04 & \\ 

385853 & OPI-JSA & OPI-JSA \cite{spiewak-sobecki-karas:2017:SemEval} &  &  &  &  &  &  & 78.50 & \\ 

386323 & OPI-JSA & OPI-JSA \cite{spiewak-sobecki-karas:2017:SemEval} &  &  &  &  &  &  & 73.42 & \\ 

385846 & OPI-JSA & OPI-JSA \cite{spiewak-sobecki-karas:2017:SemEval} &  &  &  &  &  &  & 67.96 & \\ 

386207 & PurdueNLP & PurdueNLP \cite{lee-EtAl:2017:SemEval} &  &  &  &  &  &  & 79.28 & \\ 

386645 & PurdueNLP & PurdueNLP \cite{lee-EtAl:2017:SemEval} &  &  &  &  &  &  & 55.35 & \\ 

386855 & PurdueNLP & PurdueNLP \cite{lee-EtAl:2017:SemEval} &  &  &  &  &  &  & 53.11 & \\

388410 & nlp\_2017 (*) & QLUT \cite{meng-EtAl:2017:SemEval1}* &  &  &  &  &  &  & 64.33 & \\ 

386294 & nlp\_2017 & QLUT \cite{meng-EtAl:2017:SemEval1} &  &  &  &  &  &  & 61.55 & \\ 

388499 & nlp\_2017 (*) & QLUT \cite{meng-EtAl:2017:SemEval1}* &  &  &  &  &  &  & 49.24 & \\ 

386732 & sythello & SIGMA &  &  &  &  &  &  & 80.47 & \\ 

386621 & sythello & SIGMA &  &  &  &  &  &  & 80.08 & \\

385689 & sythello & SIGMA &  &  &  &  &  &  & 79.12 & \\ 

386736 & zhangzilu919 & SIGMA\_PKU\_2 &  &  &  &  &  &  & 81.34 & \\ 

386739 & zhangzilu919 & SIGMA\_PKU\_2 &  &  &  &  &  &  & 81.27 & \\

386738 & zhangzilu919 & SIGMA\_PKU\_2 &  &  &  &  &  &  & 80.61 & \\

386823 & rekaby & STS-UHH \cite{kohail-salama-biemann:2017:SemEval} &  &  &  &  &  &  & 80.93 & \\

386377 & UCSC-NLP & UCSC-NLP &  &  &  &  &  &  & 77.29 & \\

386828 & natsheh & UdL \cite{alnatsheh-EtAl:2017:SemEval} &  &  &  &  &  &  & 80.04 & \\ 

388040 & natsheh (*) & UdL \cite{alnatsheh-EtAl:2017:SemEval}* &  &  &  &  &  &  & 79.01 & \\ 

386833 & natsheh & UdL \cite{alnatsheh-EtAl:2017:SemEval} &  &  &  &  &  &  & 78.05 & \\ 
\hline
\hline
\rowcolor{verylightgrey}
 & sts-organizers & cosine baseline & 53.70 & 60.45 & 51.55 & 71.17 & 62.20 & 3.20  & 72.78 & 54.56 \\ 
\hline
\multicolumn{11}{r}{* Corrected or late submission} \\
\end{tabu}
\end{center}
\caption{STS 2017 rankings ordered by average correlation across tracks 1-5. Performance is reported by convention as Pearson's $r \times 100$. For tracks 1-6, the top ranking result is marked with a $\bullet$ symbol and results in bold have no statistically significant difference with the best result on a track, $p > 0.05$ Williams' t-test \cite{Diedenhofen2015}.}
\label{tab:ranking-summary}
\end{table*}

\subsection{Rankings}

Participant performance is provided in Table \ref{tab:ranking-summary}. ECNU is best overall (avg r: 0.7316) and achieves the highest participant evaluation score on: track 2, Arabic-English (r: 0.7493); track 3, Spanish (r: 0.8559); and track 6, Turkish-English (r: 0.7706). BIT attains the best performance on track 1, Arabic (r: 0.7543).  CompiLIG places first on track 4a, SNLI Spanish-English (r: 0.8302). SEF@UHH exhibits the best correlation on the difficult track 4b WMT quality estimation pairs (r: 0.3407). RTV has the best system for the track 5 English data (r: 0.8547), followed closely by DT\_Team (r: 0.8536).

Especially challenging tracks with SNLI data are: track 1, Arabic; track 2, Arabic-English; and track 6, English-Turkish.  Spanish-English performance is much higher on track 4a's SNLI data than track 4b's MT quality estimation data. This highlights the difficulty and importance of making fine grained distinctions for certain downstream applications. Assessing STS  methods for quality estimation may benefit from using alternatives to Pearson correlation for evaluation.\footnote{e.g., \newcite{reimers-beyer-gurevych:2016:COLING} report success using STS labels with alternative metrics such as normalized Cumulative Gain (nCG), normalized Discounted Cumulative Gain (nDCG) and F1 to more accurately predict performance on the downstream tasks: text reuse detection, binary classification of document relatedness and document relatedness within a corpus.}

Results tend to decrease on cross-lingual tracks. The baseline drops $>10\%$ relative on Arabic-English and Spanish-English (SNLI) vs. monolingual Arabic and Spanish. Many participant systems show smaller decreases. ECNU's top ranking entry performs slightly better on Arabic-English than Arabic, with a slight drop from Spanish to Spanish-English (SNLI).

\subsection{Methods}

Participating teams explore techniques ranging from state-of-the-art deep learning models to elaborate feature engineered systems. Prediction signals include surface similarity scores such as edit distance and matching n-grams, scores derived from word alignments across pairs, assessment by MT evaluation metrics, estimates of conceptual similarity as well as the similarity between word and sentence level embeddings. For cross-lingual and non-English tracks, MT was widely used to convert the two sentences being compared into the same language.\footnote{Within the highlighted submissions, the following use a monolingual English system fed by MT: ECNU, BIT, HCTI and MITRE. HCTI submitted a separate run using ar, es and en trained models that underperformed using their en model with MT for ar and es. CompiLIG's model is cross-lingual but includes a word alignment feature that depends on MT. SEF@UHH built ar, es, en and tr models and use MT for the cross-lingual pairs. LIM-LIG and DT\_Team only participate in monolingual tracks.} Select methods are highlighted below.

\paragraph{ECNU \textnormal{\cite{tian-EtAl:2017:SemEval}}} The best overall system is from ENCU and ensembles well performing a feature engineered models with deep learning methods. Three feature engineered models use Random Forest (RF), Gradient Boosting (GB) and XGBoost (XGB) regression methods with features based on: n-gram overlap; edit distance; longest common prefix/suffix/substring; tree kernels \cite{Moschitti:2006}; word alignments \cite{Sultan2015}; summarization and MT evaluation metrics (BLEU, GTM-3, NIST, WER, METEOR, ROUGE); and kernel similarity of bags-of-words, bags-of-dependencies and pooled word-embeddings. ECNU's deep learning models are differentiated by their approach to sentence embeddings using either: averaged word embeddings, projected word embeddings, a deep averaging network (DAN) \cite{iyyer-EtAl:2015:ACL-IJCNLP} or LSTM \cite{HochreiterSchmidhuber1997}. Each network feeds the element-wise multiplication, subtraction and concatenation of paired sentence embeddings to additional layers to predict similarity scores. The ensemble averages scores from the four deep learning and three feature engineered models.\footnote{The two remaining ECNU runs only use either RF or GB and exclude the deep learning models.}

\paragraph{BIT \textnormal{\cite{wu-EtAl:2017:SemEval1}}} Second place overall is achieved by BIT primarily using sentence information content (IC) informed by WordNet and BNC word frequencies. One submission uses sentence IC exclusively. Another ensembles IC with \newcite{Sultan2015}'s alignment method, while a third ensembles IC with cosine similarity of summed word embeddings with an IDF weighting scheme. Sentence IC in isolation outperforms all systems except those from ECNU. Combining sentence IC with word embedding similarity performs best.

\paragraph{HCTI \textnormal{\cite{shao:2017:SemEval}}} Third place overall is obtained by HCTI with a model similar to a convolutional Deep Structured Semantic Model (CDSSM) \cite{chenyunnug2015,Huang2013}. Sentence embeddings are generated with twin convolutional neural networks (CNNs). The embeddings are then compared using cosine similarity and element-wise difference with the resulting values fed to additional layers to predict similarity labels. The architecture is abstractly similar to ECNU's deep learning models. UMDeep \cite{barrow-peskov:2017:SemEval} took a similar approach using LSTMs rather than CNNs for the sentence embeddings. 

\paragraph{MITRE \textnormal{\cite{henderson-EtAl:2017:SemEval}}} Fourth place overall is MITRE that, like ECNU, takes an ambitious feature engineering approach complemented by deep learning. Ensembled components include: alignment similarity; TakeLab STS \cite{saric2012takelab}; string similarity measures such as matching n-grams, summarization and MT metrics (BLEU, WER, PER, ROUGE); a RNN and recurrent convolutional neural networks (RCNN) over word alignments; and a BiLSTM that is state-of-the-art for textual entailment \cite{qianchen2016}.

\paragraph{FCICU \textnormal{\cite{hassan-EtAl:2017:SemEval}}} Fifth place overall is FCICU that computes a sense-base alignment using BabelNet \cite{Navigli:2010:BBV:1858681.1858704}. BabelNet synsets are multilingual allowing non-English and cross-lingual pairs to be processed similarly to English pairs. Alignment similarity scores are used with two runs: one that combines the scores within a string kernel and another that uses them with a weighted variant of \newcite{Sultan2015}'s method. Both runs average the Babelnet based scores with soft-cardinality \cite{JimenezSergio2012}.

\paragraph{CompiLIG \textnormal{\cite{ferrero-EtAl:2017:SemEval}}} The best Spanish-English performance on SNLI sentences was achieved by CompiLIG using the following cross-lingual features: conceptual similarity using DBNary \cite{Serasset2015}, MultiVec word embeddings \cite{AlexandreBerard2016} and character n-grams. MT is used to incorporate a similarity score based on \newcite{Brychcin2016}'s improvements to \newcite{Sultan2015}'s method.

\paragraph{LIM-LIG \textnormal{\cite{nagoudi-ferrero-schwab:2017:SemEval}}} Using only weighted word embeddings, LIM-LIG took second place on Arabic.\footnote{The approach is similar to SIF \cite{sif2017} but without removal of the common principle component} Arabic word embeddings are summed into sentence embeddings using uniform, POS and IDF weighting schemes. Sentence similarity is computed by cosine similarity. POS and IDF outperform uniform weighting. Combining the IDF and POS weights by multiplication is reported by LIM-LIG to achieve $r$ 0.7667, higher than all submitted Arabic (track 1) systems. 

\paragraph{DT\_Team \textnormal{\cite{maharjan-EtAl:2017:SemEval}}} Second place on English (track 5)\footnote{RTV took first place on track 5, English, but submitted no system description paper.} is DT\_Team using feature engineering combined with the following deep learning models: DSSM \cite{Huang2013}, CDSSM \cite{YelongShen2014} and skip-thoughts \cite{RyanKiros2015}.  Engineered features include: unigram overlap, summed word alignments scores, fraction of unaligned words, difference in word counts by type (all, adj, adverbs, nouns, verbs), and min to max ratios of words by type. Select features have a multiplicative penalty for unaligned words. 

\paragraph{SEF@UHH \textnormal{\cite{duma-menzel:2017:SemEval}}} First place on the challenging Spanish-English MT pairs (Track 4b) is SEF@UHH. Paragraph vector models \cite{quocle2014} are trained for Arabic, English, Spanish and Turkish. MT converts cross-lingual pairs into a single language and similarity scores are computed using cosine or the negation of Bray-Curtis dissimilarity. The best performing submission on track 4b uses cosine similarity of Spanish paragraph vectors with MT converting paired English sentences into Spanish.\footnote{For the cross-lingual tracks with language pair L$_1$-L$_2$, \newcite{duma-menzel:2017:SemEval} report additional experiments that vary the language choice for the paragraph vector model, using either L$_1$ or L$_2$. Experimental results are also provided that average the scores from the L$_1$ and L$_2$ models as well as that use vector correlation to compute similarity.}

\begin{table}[t]
\small
\begin{center}
\begin{tabular}{|l|r|r|r|r|} 
\hline
     Genre & Train  & Dev &  Test &Total\\
\hline
  news     &3299  &500  &500  &4299\\
  caption  &2000  &625  &525  &3250\\
  forum &    450  &375  &254  &1079\\
\hline
  total    &5749 &1500 &1379  &8628\\
\hline
\end{tabular}
\end{center}
\caption{STS Benchmark annotated examples by genres (rows) and by train, dev. test splits (columns).
}
\label{tab:stsbenchmark}
\end{table}

\begin{figure*}[h]
\small
     \subfloat[Track 5: English\label{fig:modelvshuman:en}]{
       \includegraphics[width=5cm]{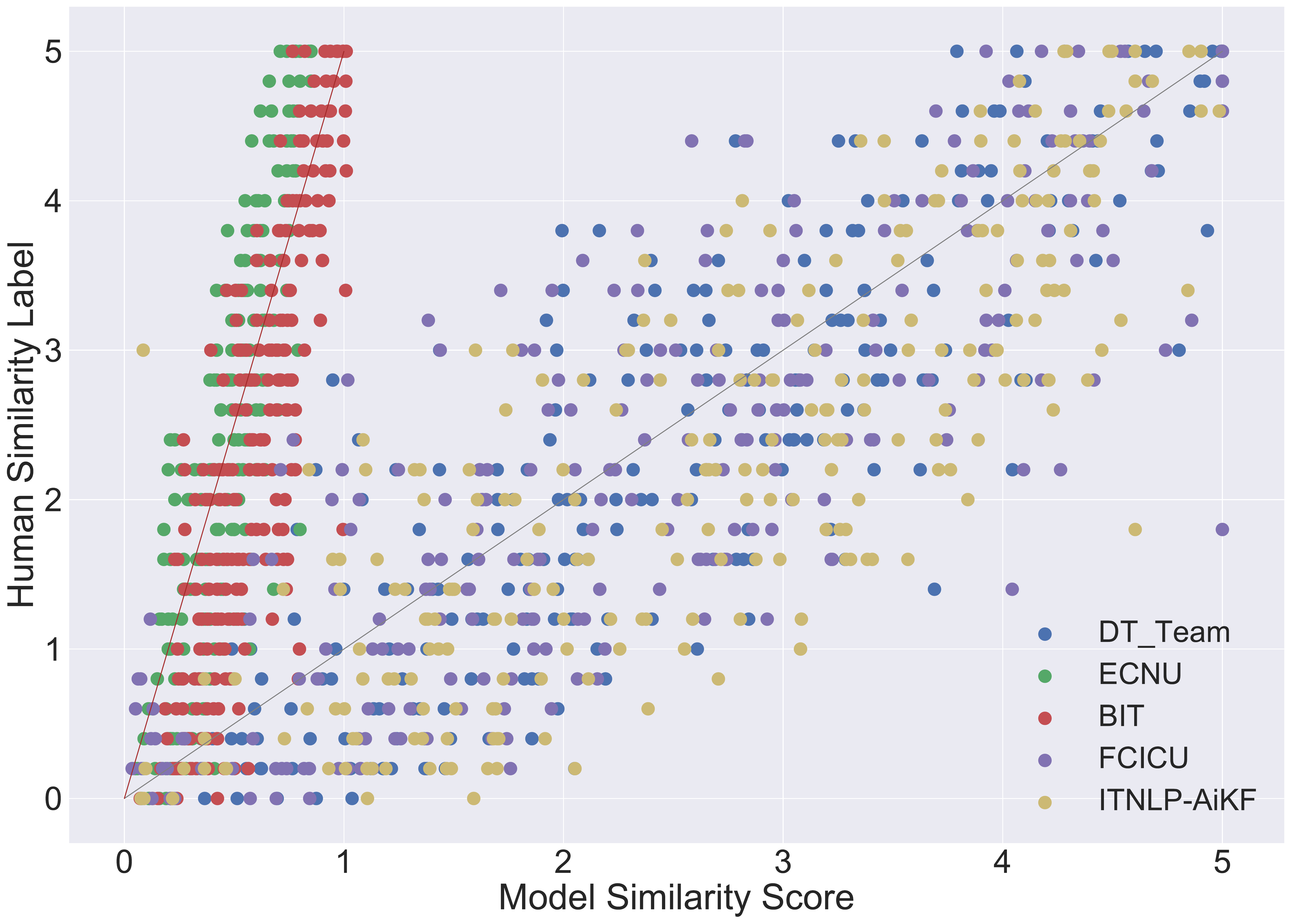}
     }
     \hfill
     \subfloat[Track 1: Arabic\label{fig:modelvshuman:ar}]{
       \includegraphics[width=5cm]{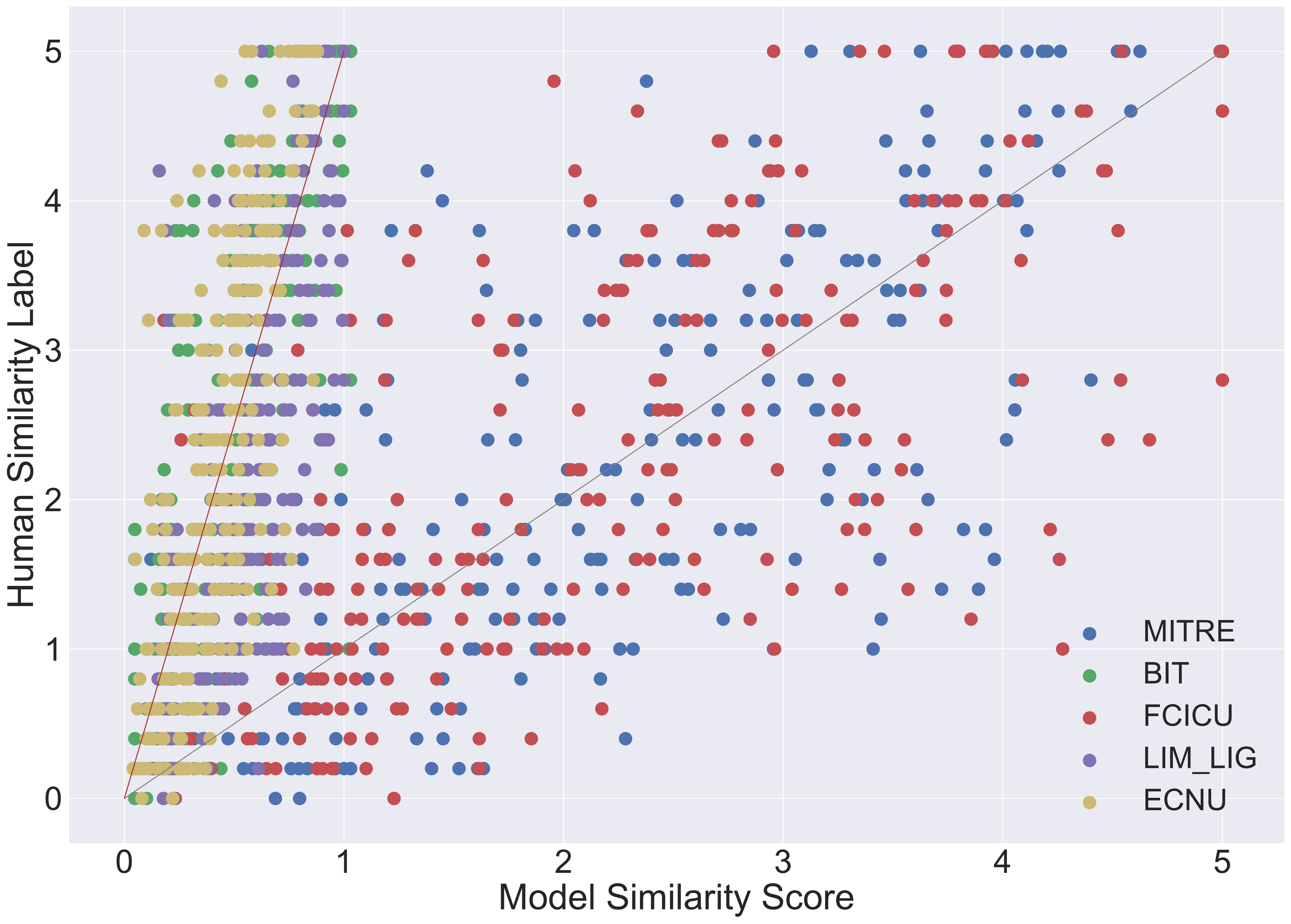}

     }
     \hfill
     \subfloat[Track 4b: Spanish-English MT\label{fig:modelvshuman:en-es}]{
       \includegraphics[width=5cm]{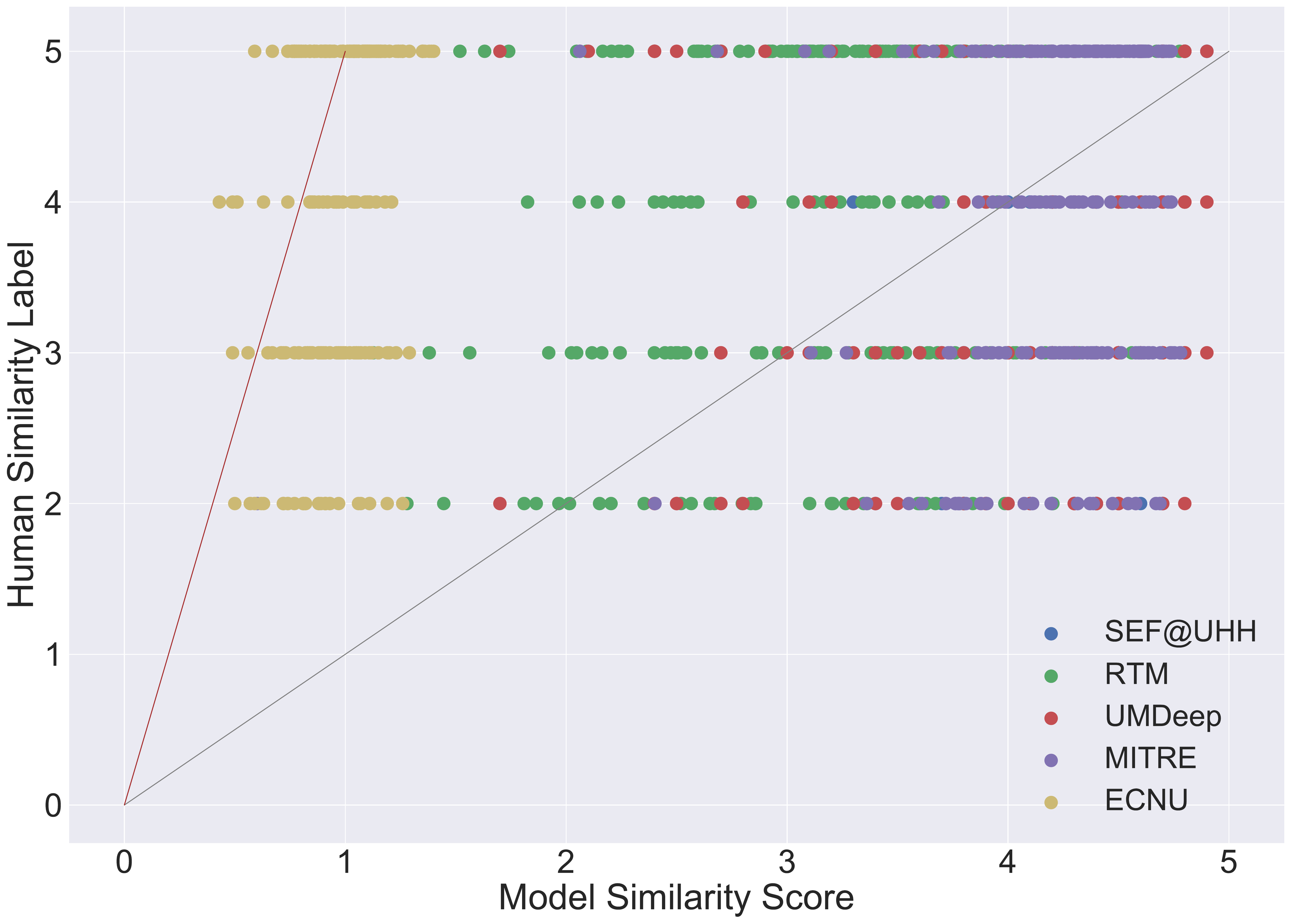}
     }
     
\caption{Model vs. human similarity scores for top systems.}
\label{fig:modelvshuman}
\end{figure*}

\begin{table*}[htp!!]
\small
\begin{center}
\centering
\begin{tabular}{|l|c|c|c|c|c|c|}
\hline
Pairs  & {\footnotesize Human} & 
{\footnotesize DT\_Team }& {\footnotesize ECNU }& {\footnotesize BIT} &  {\footnotesize FCICU }& {\footnotesize ITNLP-AiKF } \\
\hline
There is a cook preparing food. &  5.0 & 4.1 & 4.1 & 3.7 & 3.9 & 4.5 \\
A cook is making food.  & & & & & & \\
\hline
The man is in a deserted field.  & 4.0 & 3.0 & 3.1 & 3.6 & 3.1 & 2.8 \\
The man is outside in the field. & & & & & & \\ 

\hline
A girl in water without goggles or a swimming cap. & 3.0 & 4.8 & 4.6 & 4.0 & 4.7 & 0.1 \\
A girl in water, with goggles and swimming cap. & & & & & & \\
\hline
A man is carrying a canoe with a dog.  & 1.8 & 3.2 & 4.7 & 4.9 & 5.0 & 4.6 \\
A dog is carrying a man in a canoe. & & & & & & \\
\hline
There is a young girl. & 1.0 & 2.6 & 3.3 & 3.9 & 1.9 & 3.1 \\
There is a young boy with the woman. & & & & & & \\
\hline
The kids are at the theater watching a movie. & 0.2 & 1.0 & 2.3 & 2.0 & 0.8 & 1.7  \\
it is picture day for the boys & & & & & & \\
\hline
\end{tabular}
\end{center}
\caption[Example sentence pairs]{Difficult English sentence pairs (Track 5) and scores assigned by top performing systems.\footnotemark}
\label{tab:difficultpairs}
\end{table*}
\footnotetext{ECNU, BIT and LIM-LIG are scaled to the range 0-5.}
 
\begin{table}[t]
\small
\begin{center}
\begin{tabular}{|l|l|r|r|r|r|} 
\hline
 Genre     &File           &Yr.   &Train  &Dev &Test\\\hline
 news      &MSRpar         &12     &1000  &250  &250\\
 news      &headlines      &13/6  &1999  &250  &250 \\
 news      &deft-news      &14     & 300  &  0  &  0\\
 captions  &MSRvid         &12     &1000  &250  &250\\
 captions  &images         &14/5  &1000  &250  &250\\
 captions  &track5.en-en   &17      &  0  &125  &125\\
 forum     &deft-forum     &14      &450  &  0  &  0\\
 forum     &ans-forums &15      &  0  &375  &  0\\
 forum     &ans-ans  &16      &  0  &  0  &254\\
 \hline
\end{tabular}
\end{center}
\caption{STS Benchmark detailed break-down by files and years.
}
\label{tab:stsbenchmark2}
\end{table}

\begin{table*}[t]
\small
\begin{center}
\begin{tabular}{|l|l|r|r|} 
\hline
\multicolumn{4}{|c|}{STS 2017 Participants on STS Benchmark} \\
\hline
Name & Description & 	Dev &	Test\\
\hline
ECNU & 
Ensembles well performing feature eng.\@ models with deep neural networks each using sent.\@
& 84.7 	&81.0\\
    & emb.\@ from either LSTM, DAN, prj.\@ word emb.\@ or avg.\@ word emb.\@ \cite{tian-EtAl:2017:SemEval}  & & \\
BIT & Ensembles sent.\@ information content (IC) with cosine of sent.\@ emb.\@ derived from summed & 82.9 &80.9\\
    & word emb.\@ with IDF weighting scheme \cite{wu-EtAl:2017:SemEval1} & & \\
DT\_TEAM & Ensembles feature eng.\@ and deep learning signals using sent.\@ emb.\@ from DSSM, CDSSM & 83.0 	&79.2\\
 & and skip-thought models \cite{maharjan-EtAl:2017:SemEval} & & \\
UdL & Feature eng.\@ model using cosine of tf-idf weighted char n-grams, num.\@ match, sent.\@ length & 72.4 & 79.0 \\
 &  and avg. word emb.\@ cosine over PoS and NER based alignments  \cite{alnatsheh-EtAl:2017:SemEval} & &  \\
HCTI &  Deep learning model with sent.\@ emb.\@ computed using paired convolutional neural networks & 83.4 	&78.4\\
 & (CNN) and then compared using fully connected layers \cite{shao:2017:SemEval} & & \\
RTM & Referential translation machines (RTM) use a feature eng. model with transductive learning &  73.2\rlap{$^*$} & 70.6 \\
 & and parallel feature decay algorithm (ParFDA) training instance selection  \cite{biccici:2017:SemEval,Bicici:RTM:WMT2017} & & \\
SEF@UHH & Cosine of paragraph vector (PV-DBOW) sent.\@ emb.\@ \cite{duma-menzel:2017:SemEval}   & 61.6 	&59.2 \\
\hline
\multicolumn{4}{|c|}{Sentence Level Baselines} \\
\hline
InferSent & Sent.\@ emb.\@ from bi-directional LSTM trained on SNLI
\cite{Conneau2017} & 80.1 & 75.8 \\
Sent2Vec & Word \& bigram emb.\@ sum from sent.\@ spanning CBOW \cite{pgj2017unsup} & 78.7 & 75.5\\
SIF & Weighted word emb.\@ sum with principle component removal \cite{sif2017} & 80.1 & 72.0 \\
PV-DBOW & Paragraph vectors (PV-DBOW) \cite{quocle2014,Lau2016} & 72.2 & 64.9 \\
C-PHRASE & Word emb.\@ sum from model of syntactic constituent context words \cite{pham2015} & 74.3 & 63.9 \\
\hline
\multicolumn{4}{|c|}{Averaged Word Embedding Baselines} \\
\hline
LexVec & Weighted matrix factorization of PPMI \cite{Salle2016arxiv,Salle2016acl} & 68.9 & 55.8 \\
FastText & Skip-gram with sub-word character n-grams \cite{Joulin2016} & 65.2 & 53.9 \\
Paragram & Paraphrase Database (PPDB) fit word embeddings \cite{Wieting2015tacl} & 63.0 & 50.1 \\
GloVe & Word co-occurrence count fit embeddings \cite{pennington2014glove} & 52.4 & 40.6\\
Word2vec & Skip-gram prediction of words in a context window \cite{mikolov2013arxiv,mikolov2013nips} & 70.0 & 56.5 \\
\hline
\multicolumn{4}{r}{\footnotesize * 10-fold cross-validation on combination of dev and training data.} \\
\end{tabular}
\end{center}
\caption{STS Benchmark. Pearson's $r \times 100$ results for select participants and baseline models.}
\label{tab:stsbenchmarkresults}
\end{table*}
\section{Analysis}

Figure \ref{fig:modelvshuman} plots model similarity scores against human STS labels for the top 5 systems from tracks 5 (English), 1 (Arabic) and 4b (Spanish-English MT). While many systems return scores on the same scale as the gold labels, 0-5, others return scores from approximately 0 and 1. Lines on the graphs illustrate perfect performance for both a 0-5 and a 0-1 scale. Mapping the 0 to 1 scores to range from 0-5,\footnote{$s_{new} = 5 \times \frac{s-min(s)}{max(s)-min(s)}$ is used to rescale scores.} approximately 80\% of the scores from top performing English systems are within 1.0 pt of the gold label. Errors for Arabic are more broadly distributed, particularly for model scores between 1 and 4. The Spanish-English MT plots the weak relationship between the predicted and gold scores. 

Table \ref{tab:difficultpairs} provides examples of difficult sentence pairs for participant systems and illustrates common sources of error for even well-ranking systems including: (i) {\it word sense disambiguation} ``making" and ``preparing" are very similar in the context of ``food", while ``picture" and ``movie" are not similar when picture is followed by ``day";  (ii) {\it attribute importance} ``outside" vs.\ ``deserted" are smaller details when contrasting ``The man is in a deserted field" with ``The man is outside in the field"; (iii) {\it compositional meaning} ``A man is carrying a canoe with a dog" has the same content words as ``A dog is carrying a man in a canoe" but carries a different meaning; (iv) {\it negation} systems score ``\dots with goggles and a swimming cap" as nearly equivalent to ``\dots without goggles or a swimming cap". Inflated similarity scores for examples like ``There is a young girl" vs. ``There is a young boy with the woman" demonstrate (v) {\it semantic blending}, whereby appending ``with a woman" to ``boy" brings its representation closer to that of ``girl". 

For multilingual and cross-lingual pairs, these issues are magnified by translation errors for systems that use MT followed by the application of a monolingual similarity model. For  track 4b Spanish-English MT pairs, some of the poor performance can in part be attributed to many systems using MT to re-translate the output of another MT system, obscuring errors in the original translation.

\subsection{Contrasting Cross-lingual STS with MT Quality Estimation}

Since MT quality estimation pairs are translations of the same sentence, they are expected to be minimally on the same topic and have an STS score $\geq 1$.\footnote{The evaluation data for track 4b does in fact have STS scores that are $\geq 1$ for all pairs. In the 1,000 sentence training set for this track,  one sentence that received a score of zero.} The actual distribution of STS scores is such that only 13\% of the test instances score below 3, 22\% of the instances score 3, 12\% score 4 and 53\% score 5. The high STS scores indicate that MT systems are surprisingly good at preserving meaning. However, even for a human, interpreting changes caused by translations errors can be difficult due both to disfluencies and subtle errors with important changes in meaning.

The Pearson correlation between the gold MT quality scores and the gold STS scores is 0.41, which shows that translation quality measures and STS are only moderately correlated. Differences are in part explained by translation quality scores penalizing all mismatches between the source segment and its translation, whereas STS focuses on differences in meaning. However, the difficult interpretation work required for STS annotation may increase the risk of inconsistent and subjective labels. The annotations for MT quality estimation are produced as by-product of post-editing. Humans fix MT output and the edit distance between the output and its post-edited correction provides the quality score. This post-editing based procedure is known to produce relatively consistent estimates across annotators.

\section{STS Benchmark}
\label{sec:STSbenchmark}

The STS Benchmark is a careful selection of the English data sets used in SemEval and *SEM STS shared tasks between 2012 and 2017. Tables \ref{tab:stsbenchmark}  and \ref{tab:stsbenchmark2} provide details on the composition of the benchmark. 
The data is partitioned into  training, development and test sets.\footnote{Similar to the STS shared task, while the training set is provided as a convenience, researchers are encourage to incorporate other supervised and unsupervised data as long as no supervised annotations of the test partitions are used.} The development set can 
be used to design new models and tune hyperparameters. The test set should be used sparingly and only after a model design and hyperparameters have been locked against further changes. Using the STS Benchmark enables comparable assessments across different research efforts and improved tracking of the state-of-the-art.

Table \ref{tab:stsbenchmarkresults} shows the STS Benchmark results for some of the best systems from Track 5 (EN-EN)\footnote{Each participant submitted the run which did best in the development set of the STS Benchmark, which happened to be the same as their best run in Track 5 in all cases.} and compares their performance to competitive baselines from the literature. All baselines were run by the organizers using canonical pre-trained models made available by the originator of each method,\footnote{{\bf Sent2Vec}: \url{https://github.com/epfml/sent2vec}, trained model Sent2Vec\_twitter\_unigrams; {\bf SIF}: \url{https://github.com/epfml/sent2vec} Wikipedia trained word frequencies enwiki\_vocab\_min200.txt, \url{https://github.com/alexandres/lexvec} embeddings from lexvec.commoncrawl.300d.W+C.pos.vectors, first 15 principle components removed, $\alpha = 0.001$, dev experiments varied $\alpha$, principle components removed and whether GloVe, LexVec, or Word2Vec word embeddings were used; {\bf C-PHRASE}: \url{http://clic.cimec.unitn.it/composes/cphrase-vectors.html}; {\bf PV-DBOW}: \url{https://github.com/jhlau/doc2vec}, {\sc AP-NEWS} trained apnews\_dbow.tgz; {\bf LexVec}: \url{https://github.com/alexandres/lexvec}, embedddings lexvec.commoncrawl.300d.W.pos.vectors.gz; {\bf FastText}: \url{https://github.com/facebookresearch/fastText/blob/master/pretrained-vectors.md}, Wikipedia trained embeddings from wiki.en.vec; {\bf Paragram}: \url{http://ttic.uchicago.edu/~wieting/}, embeddings trained on PPDB and tuned to WS353 from Paragram-WS353; {\bf GloVe}: \url{https://nlp.stanford.edu/projects/glove/}, Wikipedia and Gigaword trained 300 dim. embeddings from glove.6B.zip; {\bf Word2vec}: \url{https://code.google.com/archive/p/word2vec/}, Google News trained embeddings from GoogleNews-vectors-negative300.bin.gz.} with the exception of PV-DBOW that uses the model from \newcite{Lau2016}  and InferSent which was reported independently. When multiple pre-trained models are available for a method, we report results for the one with the best dev set performance. For each method, input sentences are preprocessed to closely match the tokenization of the pre-trained models.\footnote{{\bf Sent2Vec}: results shown here tokenized by tweetTokenize.py constrasting dev experiments used wikiTokenize.py, both distributed with Sent2Vec. {\bf LexVec}: numbers were converted into words, all punctuation was removed, and text is lowercased; {\bf FastText}: sentences are prepared using the \texttt{normalize\_text()} function within FastText's \texttt{get-wikimedia.sh} script and lowercased; {\bf Paragram}: Joshua \cite{post2015} pipeline to pre-process and tokenized English text; {\bf C-PHRASE}, {\bf GloVe}, {\bf PV-DBOW} \& {\bf SIF}: PTB tokenization provided by Stanford CoreNLP \cite{manning-EtAl:2014:P14-5} with post-processing based on dev OOVs; {\bf Word2vec}: Similar to FastText, to our knownledge, the preprocessing for the pre-trained Word2vec embeddings is not publicly described. We use the following heuristics for the Word2vec experiment: All numbers longer than a single digit are converted into a `\#' (e.g., 24 $\rightarrow$ \#\#) then prefixed, suffixed and infixed punctuation is recursively removed from each token that does not match an entry in the model's  lexicon.} Default inference hyperparameters are used unless noted otherwise. The {\it averaged word embedding baselines} compute a sentence embedding by averaging word embeddings and then using cosine to compute pairwise sentence similarity scores.

While state-of-the-art baselines for obtaining sentence embeddings perform reasonably well on the benchmark data, improved performance is obtained by top 2017 STS shared task systems. There is still substantial room for further improvement. To follow the current state-of-the-art, visit the leaderboard on the STS wiki.\footnote{\url{http://ixa2.si.ehu.es/stswiki/index.php/STSbenchmark}}

\section{Conclusion}

We have presented the results of the 2017 STS shared task. This year's shared task differed substantially from previous iterations of STS in that the primary emphasis of the task shifted from English to multilingual and cross-lingual STS involving four different languages: Arabic, Spanish, English and Turkish. Even with this substantial change relative to prior evaluations, the shared task obtained strong participation. 31 teams produced 84 system submissions with 17 teams producing a total of 44 system submissions that processed pairs in all of the STS 2017 languages. For languages that were part of prior STS evaluations (e.g., English and Spanish), state-of-the-art systems are able to achieve strong correlations with human judgment. However, we obtain weaker correlations from participating systems for Arabic, Arabic-English and Turkish-English. This suggests further research is necessary in order to develop robust models that can both be readily applied to new languages and perform well even when less supervised training data is available. To provide a standard benchmark for English STS, we present the STS Benchmark, a careful selection of the English data sets from previous STS tasks (2012-2017). To assist in interpreting the results from new models, a number of competitive baselines and select participant systems are evaluated on the benchmark data. Ongoing improvements to the current state-of-the-art is available from an online leaderboard. 
 
\section*{Acknowledgments}
\small
We thank Alexis Conneau for the evaluation of InferSent on the STS Benchmark. This material is based in part upon work supported by QNRF-NPRP 6 - 1020-1-199 OPTDIAC that funded Arabic translation, and  by a grant from the Spanish MINECO (projects TUNER TIN2015-65308-C5-1-R and MUSTER PCIN-2015-226 cofunded by EU FEDER) that funded STS label annotation and by the QT21 EU project (H2020 No. 645452) that funded STS labels and data preparation for machine translation pairs. {I\~nigo Lopez-Gazpio is supported by the Spanish MECD. Any opinions, findings, and conclusions or recommendations expressed in this material are those of the authors and do not necessarily reflect the views of QNRF-NPRP, Spanish MINECO, QT21 EU, or the Spanish MECD.

\bibliography{acl2017}

\interlinepenalty=10000
\begin{thebibliography}{}
\expandafter\ifx\csname natexlab\endcsname\relax\def\natexlab#1{#1}\fi

\bibitem[{Agirre et~al.(2015)Agirre, Banea, Cardie, Cer, Diab, Gonzalez-Agirre,
  Guo, Lopez-Gazpio, Maritxalar, Mihalcea, Rigau, Uria, and
  Wiebe}]{agirre-EtAl:2015:SemEval}
Eneko Agirre, Carmen Banea, Claire Cardie, Daniel Cer, Mona Diab, Aitor
  Gonzalez-Agirre, Weiwei Guo, I\~{n}igo Lopez-Gazpio, Montse Maritxalar, Rada
  Mihalcea, German Rigau, Larraitz Uria, and Janyce Wiebe. 2015.
\newblock \href{http://www.aclweb.org/anthology/S15-2045}{{SemEval-2015 Task 2:
  Semantic Textual Similarity, English, Spanish and Pilot on
  Interpretability}}.
\newblock In {\em Proceedings of SemEval 2015\/}.
\newblock
  \href{http://www.aclweb.org/anthology/S15-2045}{http://www.aclweb.org/anthology/S15-2045}.

\bibitem[{Agirre et~al.(2014)Agirre, Banea, Cardie, Cer, Diab, Gonzalez-Agirre,
  Guo, Mihalcea, Rigau, and Wiebe}]{agirre-EtAl:2014:SemEval}
Eneko Agirre, Carmen Banea, Claire Cardie, Daniel Cer, Mona Diab, Aitor
  Gonzalez-Agirre, Weiwei Guo, Rada Mihalcea, German Rigau, and Janyce Wiebe.
  2014.
\newblock \href{http://www.aclweb.org/anthology/S14-2010}{{SemEval-2014 Task
  10: Multilingual semantic textual similarity}}.
\newblock In {\em Proceedings of SemEval 2014\/}.
\newblock
  \href{http://www.aclweb.org/anthology/S14-2010}{http://www.aclweb.org/anthology/S14-2010}.

\bibitem[{Agirre et~al.(2016)Agirre, Banea, Cer, Diab, Gonzalez-Agirre,
  Mihalcea, Rigau, and Wiebe}]{agirre-EtAl:2016:SemEval1}
Eneko Agirre, Carmen Banea, Daniel Cer, Mona Diab, Aitor Gonzalez-Agirre, Rada
  Mihalcea, German Rigau, and Janyce Wiebe. 2016.
\newblock \href{http://www.aclweb.org/anthology/S16-1081}{Semeval-2016 task 1:
  Semantic textual similarity, monolingual and cross-lingual evaluation}.
\newblock In {\em Proceedings of the SemEval-2016\/}.
\newblock
  \href{http://www.aclweb.org/anthology/S16-1081}{http://www.aclweb.org/anthology/S16-1081}.

\bibitem[{Agirre et~al.(2012)Agirre, Cer, Diab, and
  Gonzalez-Agirre}]{agirre-EtAl:2012:STARSEM-SEMEVAL}
Eneko Agirre, Daniel Cer, Mona Diab, and Aitor Gonzalez-Agirre. 2012.
\newblock \href{http://www.aclweb.org/anthology/S12-1051}{{SemEval-2012 Task 6:
  A pilot on semantic textual similarity}}.
\newblock In {\em Proceedings of {*SEM 2012}/{SemEval 2012}\/}.
\newblock
  \href{http://www.aclweb.org/anthology/S12-1051}{http://www.aclweb.org/anthology/S12-1051}.

\bibitem[{Agirre et~al.(2013)Agirre, Cer, Diab, Gonzalez-Agirre, and
  Guo}]{agirre-EtAl:2013:*SEM1}
Eneko Agirre, Daniel Cer, Mona Diab, Aitor Gonzalez-Agirre, and Weiwei Guo.
  2013.
\newblock \href{http://www.aclweb.org/anthology/S13-1004}{{*SEM} 2013 shared
  task: {S}emantic {T}extual {S}imilarity}.
\newblock In {\em Proceedings of {*SEM} 2013\/}.
\newblock
  \href{http://www.aclweb.org/anthology/S13-1004}{http://www.aclweb.org/anthology/S13-1004}.

\bibitem[{Al-Natsheh et~al.(2017)Al-Natsheh, Martinet, Muhlenbach, and
  ZIGHED}]{alnatsheh-EtAl:2017:SemEval}
Hussein~T. Al-Natsheh, Lucie Martinet, Fabrice Muhlenbach, and
  Djamel~Abdelkader ZIGHED. 2017.
\newblock \href{http://www.aclweb.org/anthology/S17-2013}{{UdL at SemEval-2017
  Task 1}: Semantic textual similarity estimation of english sentence pairs
  using regression model over pairwise features}.
\newblock In {\em Proceedings of SemEval-2017\/}.
\newblock
  \href{http://www.aclweb.org/anthology/S17-2013}{http://www.aclweb.org/anthology/S17-2013}.

\bibitem[{Arora et~al.(2017)Arora, Liang, and Ma}]{sif2017}
Sanjeev Arora, Yingyu Liang, and Tengyu Ma. 2017.
\newblock \href{https://openreview.net/pdf?id=SyK00v5xx}{A simple but
  tough-to-beat baseline for sentence embeddings}.
\newblock In {\em Proceedings of ICLR 2017\/}.
\newblock
  \href{https://openreview.net/pdf?id=SyK00v5xx}{https://openreview.net/pdf?id=SyK00v5xx}.

\bibitem[{Arroyo-Fern\'{a}ndez and
  Meza~Ruiz(2017)}]{arroyofernandez-mezaruiz:2017:SemEval}
Ignacio Arroyo-Fern\'{a}ndez and Ivan~Vladimir Meza~Ruiz. 2017.
\newblock \href{http://www.aclweb.org/anthology/S17-2031}{{LIPN-IIMAS at
  SemEval-2017 Task 1}: Subword embeddings, attention recurrent neural networks
  and cross word alignment for semantic textual similarity}.
\newblock In {\em Proceedings of SemEval-2017\/}.
\newblock
  \href{http://www.aclweb.org/anthology/S17-2031}{http://www.aclweb.org/anthology/S17-2031}.

\bibitem[{Baker et~al.(1998)Baker, Fillmore, and Lowe}]{Baker:98}
Collin~F. Baker, Charles~J. Fillmore, and John~B. Lowe. 1998.
\newblock \href{http://aclweb.org/anthology/P/P98/P98-1013.pdf}{{The Berkeley
  FrameNet Project}}.
\newblock In {\em Proceedings of COLING '98\/}.
\newblock
  \href{http://aclweb.org/anthology/P/P98/P98-1013.pdf}{http://aclweb.org/anthology/P/P98/P98-1013.pdf}.

\bibitem[{B\"{a}r et~al.(2012)B\"{a}r, Biemann, Gurevych, and
  Zesch}]{bar-EtAl:2012:STARSEM-SEMEVAL}
Daniel B\"{a}r, Chris Biemann, Iryna Gurevych, and Torsten Zesch. 2012.
\newblock \href{http://www.aclweb.org/anthology/S12-1059}{Ukp: Computing
  semantic textual similarity by combining multiple content similarity
  measures}.
\newblock In {\em Proceedings of {*SEM 2012}/{SemEval 2012}\/}.
\newblock
  \href{http://www.aclweb.org/anthology/S12-1059}{http://www.aclweb.org/anthology/S12-1059}.

\bibitem[{Barrow and Peskov(2017)}]{barrow-peskov:2017:SemEval}
Joe Barrow and Denis Peskov. 2017.
\newblock \href{http://www.aclweb.org/anthology/S17-2026}{{UMDeep at
  SemEval-2017 Task 1}: End-to-end shared weight {LSTM} model for semantic
  textual similarity}.
\newblock In {\em Proceedings of SemEval-2017\/}.
\newblock
  \href{http://www.aclweb.org/anthology/S17-2026}{http://www.aclweb.org/anthology/S17-2026}.

\bibitem[{Bentivogli et~al.(2016)Bentivogli, Bernardi, Marelli, Menini, Baroni,
  and Zamparelli}]{Bentivogli2016}
Luisa Bentivogli, Raffaella Bernardi, Marco Marelli, Stefano Menini, Marco
  Baroni, and Roberto Zamparelli. 2016.
\newblock \href{https://doi.org/10.1007/s10579-015-9332-5}{{SICK} through the
  {SemEval} glasses. lesson learned from the evaluation of compositional
  distributional semantic models on full sentences through semantic relatedness
  and textual entailment}.
\newblock {\em Lang Resour Eval\/} 50(1):95--124.
\newblock
  \href{https://doi.org/10.1007/s10579-015-9332-5}{https://doi.org/10.1007/s10579-015-9332-5}.

\bibitem[{Berard et~al.(2016)Berard, Servan, Pietquin, and
  Besacier}]{AlexandreBerard2016}
Alexandre Berard, Christophe Servan, Olivier Pietquin, and Laurent Besacier.
  2016.
\newblock
  \href{{http://www.lrec-conf.org/proceedings/lrec2016/pdf/666\_Paper.pdf}}{{MultiVec}:
  a multilingual and multilevel representation learning toolkit for {NLP}}.
\newblock In {\em Proceedings of LREC 2016\/}.
\newblock
  \href{{http://www.lrec-conf.org/proceedings/lrec2016/pdf/666\_Paper.pdf}}{{http://www.lrec-conf.org/proceedings/lrec2016/pdf/666\_Paper.pdf}}.

\bibitem[{Bi\c{c}ici(2017{\natexlab{a}})}]{Bicici:RTM:WMT2017}
Ergun Bi\c{c}ici. 2017{\natexlab{a}}.
\newblock Predicting translation performance with referential translation
  machines.
\newblock In {\em Proceedings of WMT17 (to appear)\/}.

\bibitem[{Bi\c{c}ici(2017{\natexlab{b}})}]{biccici:2017:SemEval}
Ergun Bi\c{c}ici. 2017{\natexlab{b}}.
\newblock \href{http://www.aclweb.org/anthology/S17-2030}{{RTM at SemEval-2017
  Task 1}: Referential translation machines for predicting semantic
  similarity}.
\newblock In {\em Proceedings of SemEval-2017\/}.
\newblock
  \href{http://www.aclweb.org/anthology/S17-2030}{http://www.aclweb.org/anthology/S17-2030}.

\bibitem[{Bjerva and \"{O}stling(2017)}]{bjerva-ostling:2017:SemEval}
Johannes Bjerva and Robert \"{O}stling. 2017.
\newblock \href{http://www.aclweb.org/anthology/S17-2021}{{ResSim at
  SemEval-2017 Task 1}: Multilingual word representations for semantic textual
  similarity}.
\newblock In {\em Proceedings of SemEval-2017\/}.
\newblock
  \href{http://www.aclweb.org/anthology/S17-2021}{http://www.aclweb.org/anthology/S17-2021}.

\bibitem[{Bojar et~al.(2014)Bojar, Buck, Federmann, Haddow, Koehn, Leveling,
  Monz, Pecina, Post, Saint-Amand, Soricut, Specia, and
  Tamchyna}]{bojar-etal_WMT:2014}
Ondrej Bojar, Christian Buck, Christian Federmann, Barry Haddow, Philipp Koehn,
  Johannes Leveling, Christof Monz, Pavel Pecina, Matt Post, Herve Saint-Amand,
  Radu Soricut, Lucia Specia, and Ale\v{s} Tamchyna. 2014.
\newblock \href{http://www.aclweb.org/anthology/W/W14/W14-3302.pdf}{Findings of
  the 2014 workshop on statistical machine translation}.
\newblock In {\em Proceedings of WMT 2014\/}.
\newblock
  \href{http://www.aclweb.org/anthology/W/W14/W14-3302.pdf}{http://www.aclweb.org/anthology/W/W14/W14-3302.pdf}.

\bibitem[{Bojar et~al.(2013)Bojar, Buck, Callison-Burch, Federmann, Haddow,
  Koehn, Monz, Post, Soricut, and Specia}]{bojar-EtAl:2013:WMT}
Ond\v{r}ej Bojar, Christian Buck, Chris Callison-Burch, Christian Federmann,
  Barry Haddow, Philipp Koehn, Christof Monz, Matt Post, Radu Soricut, and
  Lucia Specia. 2013.
\newblock \href{http://www.aclweb.org/anthology/W13-2201}{Findings of the 2013
  {Workshop on Statistical Machine Translation}}.
\newblock In {\em Proceedings of WMT 2013\/}.
\newblock
  \href{http://www.aclweb.org/anthology/W13-2201}{http://www.aclweb.org/anthology/W13-2201}.

\bibitem[{Bowman et~al.(2015)Bowman, Angeli, Potts, and
  Manning}]{snli:emnlp2015}
Samuel~R. Bowman, Gabor Angeli, Christopher Potts, and Christopher~D. Manning.
  2015.
\newblock \href{http://aclweb.org/anthology/D/D15/D15-1075.pdf}{A large
  annotated corpus for learning natural language inference}.
\newblock In {\em Proceedings of EMNLP 2015\/}.
\newblock
  \href{http://aclweb.org/anthology/D/D15/D15-1075.pdf}{http://aclweb.org/anthology/D/D15/D15-1075.pdf}.

\bibitem[{Brychcin and Svoboda(2016)}]{Brychcin2016}
Tomas Brychcin and Lukas Svoboda. 2016.
\newblock \href{https://www.aclweb.org/anthology/S/S16/S16-1089.pdf}{{UWB} at
  {SemEval-2016 Task 1}: Semantic textual similarity using lexical, syntactic,
  and semantic information}.
\newblock In {\em Proceedings of SemEval 2016\/}.
\newblock
  \href{https://www.aclweb.org/anthology/S/S16/S16-1089.pdf}{https://www.aclweb.org/anthology/S/S16/S16-1089.pdf}.

\bibitem[{Chen et~al.(2016)Chen, Zhu, Ling, Wei, and Jiang}]{qianchen2016}
Qian Chen, Xiaodan Zhu, Zhen{-}Hua Ling, Si~Wei, and Hui Jiang. 2016.
\newblock \href{http://arxiv.org/abs/1609.06038}{Enhancing and combining
  sequential and tree {LSTM} for natural language inference}.
\newblock {\em CoRR\/} abs/1609.06038.
\newblock
  \href{http://arxiv.org/abs/1609.06038}{http://arxiv.org/abs/1609.06038}.

\bibitem[{Chen et~al.(2015)Chen, Hakkani-T\"{u}r, and He}]{chenyunnug2015}
Yun-Nung Chen, Dilek Hakkani-T\"{u}r, and Xiaodong He. 2015.
\newblock
  \href{https://www.microsoft.com/en-us/research/publication/learning-bidirectional-intent-embeddings-by-convolutional-deep-structured-semantic-models-for-spoken-language-understanding/}{Learning
  bidirectional intent embeddings by convolutional deep structured semantic
  models for spoken language understanding}.
\newblock In {\em Proceedings of NIPS-SLU, 2015\/}.
\newblock
  \href{https://www.microsoft.com/en-us/research/publication/learning-bidirectional-intent-embeddings-by-convolutional-deep-structured-semantic-models-for-spoken-language-understanding/}{https://www.microsoft.com/en-us/research/publication/learning-bidirectional-intent-embeddings-by-convolutional-deep-structured-semantic-models-for-spoken-language-understanding/}.

\bibitem[{Conneau et~al.(2017)Conneau, Kiela, Schwenk, Barrault, and
  Bordes}]{Conneau2017}
Alexis Conneau, Douwe Kiela, Holger Schwenk, Lo{\"{\i}}c Barrault, and Antoine
  Bordes. 2017.
\newblock \href{http://arxiv.org/abs/1705.02364}{Supervised learning of
  universal sentence representations from natural language inference data}.
\newblock {\em CoRR\/} abs/1705.02364.
\newblock
  \href{http://arxiv.org/abs/1705.02364}{http://arxiv.org/abs/1705.02364}.

\bibitem[{Dagan et~al.(2010)Dagan, Dolan, Magnini, and Roth}]{dagan2009}
Ido Dagan, Bill Dolan, Bernardo Magnini, and Dan Roth. 2010.
\newblock \href{https://doi.org/10.1017/S1351324909990234}{Recognizing textual
  entailment: Rational, evaluation and approaches}.
\newblock {\em J. Nat. Language Eng.\/} 16:105--105.
\newblock
  \href{https://doi.org/10.1017/S1351324909990234}{https://doi.org/10.1017/S1351324909990234}.

\bibitem[{Diedenhofen and Musch(2015)}]{Diedenhofen2015}
Birk Diedenhofen and Jochen Musch. 2015.
\newblock \href{http://dx.doi.org/10.1371/journal.pone.0121945}{cocor: A
  comprehensive solution for the statistical comparison of correlations}.
\newblock {\em PLoS ONE\/} 10(4).
\newblock
  \href{http://dx.doi.org/10.1371/journal.pone.0121945}{http://dx.doi.org/10.1371/journal.pone.0121945}.

\bibitem[{Dolan et~al.(2004)Dolan, Quirk, and Brockett}]{:msrpp}
Bill Dolan, Chris Quirk, and Chris Brockett. 2004.
\newblock \href{http://aclweb.org/anthology/C/C04/C04-1051.pdf}{Unsupervised
  construction of large paraphrase corpora: Exploiting massively parallel news
  sources}.
\newblock In {\em Proceedings of COLING ’04\/}.
\newblock
  \href{http://aclweb.org/anthology/C/C04/C04-1051.pdf}{http://aclweb.org/anthology/C/C04/C04-1051.pdf}.

\bibitem[{Duma and Menzel(2017)}]{duma-menzel:2017:SemEval}
Mirela-Stefania Duma and Wolfgang Menzel. 2017.
\newblock \href{http://www.aclweb.org/anthology/S17-2024}{{SEF$@$UHH at
  SemEval-2017 Task 1}: Unsupervised knowledge-free semantic textual similarity
  via paragraph vector}.
\newblock In {\em Proceedings of SemEval-2017\/}.
\newblock
  \href{http://www.aclweb.org/anthology/S17-2024}{http://www.aclweb.org/anthology/S17-2024}.

\bibitem[{Espa\~{n}a Bonet and
  Barr\'{o}n-Cede\~{n}o(2017)}]{espanabonet-barroncedeno:2017:SemEval}
Cristina Espa\~{n}a Bonet and Alberto Barr\'{o}n-Cede\~{n}o. 2017.
\newblock \href{http://www.aclweb.org/anthology/S17-2019}{{Lump at SemEval-2017
  Task 1}: Towards an interlingua semantic similarity}.
\newblock In {\em Proceedings of SemEval-2017\/}.
\newblock
  \href{http://www.aclweb.org/anthology/S17-2019}{http://www.aclweb.org/anthology/S17-2019}.

\bibitem[{Fellbaum(1998)}]{Fellbaum:98}
Christiane Fellbaum. 1998.
\newblock {\em {WordNet}: An Electronic Lexical Database\/}.
\newblock MIT Press.
\newblock
  \href{https://books.google.com/books?id=Rehu8OOzMIMC}{https://books.google.com/books?id=Rehu8OOzMIMC}.

\bibitem[{Ferrero et~al.(2017)Ferrero, Besacier, Schwab, and
  Agn\`{e}s}]{ferrero-EtAl:2017:SemEval}
J\'{e}r\'{e}my Ferrero, Laurent Besacier, Didier Schwab, and Fr\'{e}d\'{e}ric
  Agn\`{e}s. 2017.
\newblock \href{http://www.aclweb.org/anthology/S17-2012}{{CompiLIG at
  SemEval-2017 Task 1}: Cross-language plagiarism detection methods for
  semantic textual similarity}.
\newblock In {\em Proceedings of SemEval-2017\/}.
\newblock
  \href{http://www.aclweb.org/anthology/S17-2012}{http://www.aclweb.org/anthology/S17-2012}.

\bibitem[{Fialho et~al.(2017)Fialho, Patinho~Rodrigues, Coheur, and
  Quaresma}]{fialho-EtAl:2017:SemEval}
Pedro Fialho, Hugo Patinho~Rodrigues, Lu\'{i}sa Coheur, and Paulo Quaresma.
  2017.
\newblock \href{http://www.aclweb.org/anthology/S17-2032}{L2f/inesc-id at
  semeval-2017 tasks 1 and 2: Lexical and semantic features in word and textual
  similarity}.
\newblock In {\em Proceedings of SemEval-2017\/}.
\newblock
  \href{http://www.aclweb.org/anthology/S17-2032}{http://www.aclweb.org/anthology/S17-2032}.

\bibitem[{Ganitkevitch et~al.(2013)Ganitkevitch, {Van Durme}, and
  Callison-Burch}]{ganitkevitch2013ppdb}
Juri Ganitkevitch, Benjamin {Van Durme}, and Chris Callison-Burch. 2013.
\newblock \href{http://cs.jhu.edu/~ccb/publications/ppdb.pdf}{{PPDB}: The
  paraphrase database}.
\newblock In {\em Proceedings of NAACL/HLT 2013\/}.
\newblock
  \href{http://cs.jhu.edu/~ccb/publications/ppdb.pdf}{http://cs.jhu.edu/~ccb/publications/ppdb.pdf}.

\bibitem[{Hassan et~al.(2017)Hassan, AbdelRahman, Bahgat, and
  Farag}]{hassan-EtAl:2017:SemEval}
Basma Hassan, Samir AbdelRahman, Reem Bahgat, and Ibrahim Farag. 2017.
\newblock \href{http://www.aclweb.org/anthology/S17-2015}{{FCICU} at
  {SemEval-2017 Task 1}: Sense-based language independent semantic textual
  similarity approach}.
\newblock In {\em Proceedings of SemEval-2017\/}.
\newblock
  \href{http://www.aclweb.org/anthology/S17-2015}{http://www.aclweb.org/anthology/S17-2015}.

\bibitem[{He et~al.(2015)He, Gimpel, and Lin}]{he-gimpel-lin:2015:EMNLP}
Hua He, Kevin Gimpel, and Jimmy Lin. 2015.
\newblock \href{http://aclweb.org/anthology/D15-1181}{Multi-perspective
  sentence similarity modeling with convolutional neural networks}.
\newblock In {\em Proceedings of EMNLP\/}. pages 1576--1586.
\newblock
  \href{http://aclweb.org/anthology/D15-1181}{http://aclweb.org/anthology/D15-1181}.

\bibitem[{He and Lin(2016)}]{he-lin:2016:N16-1}
Hua He and Jimmy Lin. 2016.
\newblock \href{http://www.aclweb.org/anthology/N16-1108}{Pairwise word
  interaction modeling with deep neural networks for semantic similarity
  measurement}.
\newblock In {\em Proceedings of NAACL/HLT\/}.
\newblock
  \href{http://www.aclweb.org/anthology/N16-1108}{http://www.aclweb.org/anthology/N16-1108}.

\bibitem[{He et~al.(2016)He, Wieting, Gimpel, Rao, and Lin}]{He2016}
Hua He, John Wieting, Kevin Gimpel, Jinfeng Rao, and Jimmy Lin. 2016.
\newblock
  \href{http://www.anthology.aclweb.org/S/S16/S16-1170.pdf}{{UMD-TTIC-UW} at
  {SemEval-2016 Task 1}: Attention-based multi-perspective convolutional neural
  networks for textual similarity measurement}.
\newblock In {\em Proceedings of SemEval 2016\/}.
\newblock
  \href{http://www.anthology.aclweb.org/S/S16/S16-1170.pdf}{http://www.anthology.aclweb.org/S/S16/S16-1170.pdf}.

\bibitem[{Henderson et~al.(2017)Henderson, Merkhofer, Strickhart, and
  Zarrella}]{henderson-EtAl:2017:SemEval}
John Henderson, Elizabeth Merkhofer, Laura Strickhart, and Guido Zarrella.
  2017.
\newblock \href{http://www.aclweb.org/anthology/S17-2027}{{MITRE} at
  {SemEval-2017 Task 1}: Simple semantic similarity}.
\newblock In {\em Proceedings of SemEval-2017\/}.
\newblock
  \href{http://www.aclweb.org/anthology/S17-2027}{http://www.aclweb.org/anthology/S17-2027}.

\bibitem[{Hill et~al.(2016)Hill, Cho, and Korhonen}]{hill2016}
Felix Hill, Kyunghyun Cho, and Anna Korhonen. 2016.
\newblock \href{http://www.aclweb.org/anthology/N16-1162}{Learning distributed
  representations of sentences from unlabelled data}.
\newblock In {\em Proceedings of NAACL/HLT\/}.
\newblock
  \href{http://www.aclweb.org/anthology/N16-1162}{http://www.aclweb.org/anthology/N16-1162}.

\bibitem[{Hochreiter and Schmidhuber(1997)}]{HochreiterSchmidhuber1997}
Sepp Hochreiter and J\"{u}rgen Schmidhuber. 1997.
\newblock \href{http://dx.doi.org/10.1162/neco.1997.9.8.1735}{Long short-term
  memory}.
\newblock {\em Neural Comput.\/} 9(8):1735--1780.
\newblock
  \href{http://dx.doi.org/10.1162/neco.1997.9.8.1735}{http://dx.doi.org/10.1162/neco.1997.9.8.1735}.

\bibitem[{Hovy et~al.(2006)Hovy, Marcus, Palmer, Ramshaw, and
  Weischedel}]{Hovy:06}
Eduard Hovy, Mitchell Marcus, Martha Palmer, Lance Ramshaw, and Ralph
  Weischedel. 2006.
\newblock \href{http://aclweb.org/anthology/N/N06/N06-2015.pdf}{{OntoNotes}:
  The 90\% solution}.
\newblock In {\em Proceedings of NAACL/HLT 2006\/}.
\newblock
  \href{http://aclweb.org/anthology/N/N06/N06-2015.pdf}{http://aclweb.org/anthology/N/N06/N06-2015.pdf}.

\bibitem[{Huang et~al.(2013)Huang, He, Gao, Deng, Acero, and Heck}]{Huang2013}
Po-Sen Huang, Xiaodong He, Jianfeng Gao, Li~Deng, Alex Acero, and Larry Heck.
  2013.
\newblock
  \href{https://www.microsoft.com/en-us/research/publication/learning-deep-structured-semantic-models-for-web-search-using-clickthrough-data/}{Learning
  deep structured semantic models for web search using clickthrough data}.
\newblock In {\em Proceedings of CIKM\/}.
\newblock
  \href{https://www.microsoft.com/en-us/research/publication/learning-deep-structured-semantic-models-for-web-search-using-clickthrough-data/}{https://www.microsoft.com/en-us/research/publication/learning-deep-structured-semantic-models-for-web-search-using-clickthrough-data/}.

\bibitem[{Iyyer et~al.(2015)Iyyer, Manjunatha, Boyd-Graber, and
  Daum\'{e}~III}]{iyyer-EtAl:2015:ACL-IJCNLP}
Mohit Iyyer, Varun Manjunatha, Jordan Boyd-Graber, and Hal Daum\'{e}~III. 2015.
\newblock \href{http://www.aclweb.org/anthology/P15-1162}{Deep unordered
  composition rivals syntactic methods for text classification}.
\newblock In {\em Proceedings of ACL/IJCNLP\/}.
\newblock
  \href{http://www.aclweb.org/anthology/P15-1162}{http://www.aclweb.org/anthology/P15-1162}.

\bibitem[{Jimenez et~al.(2012{\natexlab{a}})Jimenez, Becerra, and
  Gelbukh}]{jimenez-becerra-gelbukh:2012:STARSEM-SEMEVAL1}
Sergio Jimenez, Claudia Becerra, and Alexander Gelbukh. 2012{\natexlab{a}}.
\newblock \href{http://www.aclweb.org/anthology/S12-1061}{Soft cardinality: A
  parameterized similarity function for text comparison}.
\newblock In {\em Proceedings of {*SEM 2012}/{SemEval 2012}\/}.
\newblock
  \href{http://www.aclweb.org/anthology/S12-1061}{http://www.aclweb.org/anthology/S12-1061}.

\bibitem[{Jimenez et~al.(2012{\natexlab{b}})Jimenez, Becerra, and
  Gelbukh}]{JimenezSergio2012}
Sergio Jimenez, Claudia Becerra, and Alexander Gelbukh. 2012{\natexlab{b}}.
\newblock \href{http://aclweb.org/anthology/S/S12/S12-1061.pdf}{{Soft
  Cardinality}: A parameterized similarity function for text comparison}.
\newblock In {\em Proceedings of *SEM 2012/SemEval 2012\/}.
\newblock
  \href{http://aclweb.org/anthology/S/S12/S12-1061.pdf}{http://aclweb.org/anthology/S/S12/S12-1061.pdf}.

\bibitem[{Joulin et~al.(2016)Joulin, Grave, Bojanowski, and
  Mikolov}]{Joulin2016}
Armand Joulin, Edouard Grave, Piotr Bojanowski, and Tomas Mikolov. 2016.
\newblock \href{http://arxiv.org/abs/1607.01759}{Bag of tricks for efficient
  text classification}.
\newblock {\em CoRR\/} abs/1607.01759.
\newblock
  \href{http://arxiv.org/abs/1607.01759}{http://arxiv.org/abs/1607.01759}.

\bibitem[{Kenter et~al.(2016)Kenter, Borisov, and de~Rijke}]{kenter2016}
Tom Kenter, Alexey Borisov, and Maarten de~Rijke. 2016.
\newblock \href{http://www.aclweb.org/anthology/P16-1089}{Siamese cbow:
  Optimizing word embeddings for sentence representations}.
\newblock In {\em Proceedings of ACL\/}.
\newblock
  \href{http://www.aclweb.org/anthology/P16-1089}{http://www.aclweb.org/anthology/P16-1089}.

\bibitem[{Kiros et~al.(2015)Kiros, Zhu, Salakhutdinov, Zemel, Torralba,
  Urtasun, and Fidler}]{RyanKiros2015}
Ryan Kiros, Yukun Zhu, Ruslan Salakhutdinov, Richard~S. Zemel, Antonio
  Torralba, Raquel Urtasun, and Sanja Fidler. 2015.
\newblock \href{http://arxiv.org/abs/1506.06726}{Skip-thought vectors}.
\newblock {\em CoRR\/} abs/1506.06726.
\newblock
  \href{http://arxiv.org/abs/1506.06726}{http://arxiv.org/abs/1506.06726}.

\bibitem[{Kohail et~al.(2017)Kohail, Salama, and
  Biemann}]{kohail-salama-biemann:2017:SemEval}
Sarah Kohail, Amr~Rekaby Salama, and Chris Biemann. 2017.
\newblock \href{http://www.aclweb.org/anthology/S17-2025}{{STS-UHH at
  SemEval-2017 Task 1}: Scoring semantic textual similarity using supervised
  and unsupervised ensemble}.
\newblock In {\em Proceedings of SemEval-2017\/}.
\newblock
  \href{http://www.aclweb.org/anthology/S17-2025}{http://www.aclweb.org/anthology/S17-2025}.

\bibitem[{Lau and Baldwin(2016)}]{Lau2016}
Jey~Han Lau and Timothy Baldwin. 2016.
\newblock \href{http://www.aclweb.org/anthology/W/W16/W16-1609.pdf}{An
  empirical evaluation of doc2vec with practical insights into document
  embedding generation}.
\newblock In {\em Proceedings of {ACL} Workshop on Representation Learning for
  {NLP}\/}.
\newblock
  \href{http://www.aclweb.org/anthology/W/W16/W16-1609.pdf}{http://www.aclweb.org/anthology/W/W16/W16-1609.pdf}.

\bibitem[{Le and Mikolov(2014)}]{quocle2014}
Quoc~V. Le and Tomas Mikolov. 2014.
\newblock \href{http://arxiv.org/abs/1405.4053}{Distributed representations of
  sentences and documents}.
\newblock {\em CoRR\/} abs/1405.4053.
\newblock
  \href{http://arxiv.org/abs/1405.4053}{http://arxiv.org/abs/1405.4053}.

\bibitem[{Lee et~al.(2017)Lee, Goindani, Li, Jin, Johnson, Zhang, Pacheco, and
  Goldwasser}]{lee-EtAl:2017:SemEval}
I-Ta Lee, Mahak Goindani, Chang Li, Di~Jin, Kristen~Marie Johnson, Xiao Zhang,
  Maria~Leonor Pacheco, and Dan Goldwasser. 2017.
\newblock \href{http://www.aclweb.org/anthology/S17-2029}{{PurdueNLP at
  SemEval-2017 Task 1}: Predicting semantic textual similarity with paraphrase
  and event embeddings}.
\newblock In {\em Proceedings of SemEval-2017\/}.
\newblock
  \href{http://www.aclweb.org/anthology/S17-2029}{http://www.aclweb.org/anthology/S17-2029}.

\bibitem[{Liu et~al.(2017)Liu, Sun, Lin, and Liu}]{liu-EtAl:2017:SemEval1}
Wenjie Liu, Chengjie Sun, Lei Lin, and Bingquan Liu. 2017.
\newblock \href{http://www.aclweb.org/anthology/S17-2022}{{ITNLP-AiKF at
  SemEval-2017 Task 1}: Rich features based svr for semantic textual similarity
  computing}.
\newblock In {\em Proceedings of SemEval-2017\/}.
\newblock
  \href{http://www.aclweb.org/anthology/S17-2022}{http://www.aclweb.org/anthology/S17-2022}.

\bibitem[{Maharjan et~al.(2017)Maharjan, Banjade, Gautam, Tamang, and
  Rus}]{maharjan-EtAl:2017:SemEval}
Nabin Maharjan, Rajendra Banjade, Dipesh Gautam, Lasang~J. Tamang, and Vasile
  Rus. 2017.
\newblock \href{http://www.aclweb.org/anthology/S17-2014}{Dt\_team at
  semeval-2017 task 1: Semantic similarity using alignments, sentence-level
  embeddings and gaussian mixture model output}.
\newblock In {\em Proceedings of SemEval-2017\/}.
\newblock
  \href{http://www.aclweb.org/anthology/S17-2014}{http://www.aclweb.org/anthology/S17-2014}.

\bibitem[{Manning et~al.(2014)Manning, Surdeanu, Bauer, Finkel, Bethard, and
  McClosky}]{manning-EtAl:2014:P14-5}
Christopher~D. Manning, Mihai Surdeanu, John Bauer, Jenny Finkel, Steven~J.
  Bethard, and David McClosky. 2014.
\newblock \href{http://www.aclweb.org/anthology/P/P14/P14-5010}{The {Stanford}
  {CoreNLP} natural language processing toolkit}.
\newblock In {\em Proceedings of ACL 2014 Demonstrations\/}.
\newblock
  \href{http://www.aclweb.org/anthology/P/P14/P14-5010}{http://www.aclweb.org/anthology/P/P14/P14-5010}.

\bibitem[{Marelli et~al.(2014)Marelli, Menini, Baroni, Bentivogli, Bernardi,
  and Zamparelli}]{MARELLI14.363}
Marco Marelli, Stefano Menini, Marco Baroni, Luisa Bentivogli, Raffaella
  Bernardi, and Roberto Zamparelli. 2014.
\newblock
  \href{http://www.lrec-conf.org/proceedings/lrec2014/pdf/363\_Paper.pdf}{A
  {SICK} cure for the evaluation of compositional distributional semantic
  models}.
\newblock In {\em Proceedings of LREC 14\/}.
\newblock
  \href{http://www.lrec-conf.org/proceedings/lrec2014/pdf/363\_Paper.pdf}{http://www.lrec-conf.org/proceedings/lrec2014/pdf/363\_Paper.pdf}.

\bibitem[{Matt~Post(2015)}]{post2015}
Yuan~Cao Gaurav~Kumar Matt~Post. 2015.
\newblock
  \href{https://ufal.mff.cuni.cz/pbml/104/art-post-cao-kumar.pdf}{Joshua 6: A
  phrase-based and hierarchical statistical machine translation}.
\newblock {\em The Prague Bulletin of Mathematical Linguistics\/} 104:5–16.
\newblock
  \href{https://ufal.mff.cuni.cz/pbml/104/art-post-cao-kumar.pdf}{https://ufal.mff.cuni.cz/pbml/104/art-post-cao-kumar.pdf}.

\bibitem[{Meng et~al.(2017)Meng, Lu, Zhang, Cheng, Du, and
  Han}]{meng-EtAl:2017:SemEval1}
Fanqing Meng, Wenpeng Lu, Yuteng Zhang, Jinyong Cheng, Yuehan Du, and Shuwang
  Han. 2017.
\newblock \href{http://www.aclweb.org/anthology/S17-2020}{{QLUT at SemEval-2017
  Task 1}: Semantic textual similarity based on word embeddings}.
\newblock In {\em Proceedings of SemEval-2017\/}.
\newblock
  \href{http://www.aclweb.org/anthology/S17-2020}{http://www.aclweb.org/anthology/S17-2020}.

\bibitem[{Mikolov et~al.(2013{\natexlab{a}})Mikolov, Chen, Corrado, and
  Dean}]{mikolov2013arxiv}
Tomas Mikolov, Kai Chen, Greg Corrado, and Jeffrey Dean. 2013{\natexlab{a}}.
\newblock \href{http://arxiv.org/abs/1301.3781}{Efficient estimation of word
  representations in vector space}.
\newblock {\em CoRR\/} abs/1301.3781.
\newblock
  \href{http://arxiv.org/abs/1301.3781}{http://arxiv.org/abs/1301.3781}.

\bibitem[{Mikolov et~al.(2013{\natexlab{b}})Mikolov, Sutskever, Chen, Corrado,
  and Dean}]{mikolov2013nips}
Tomas Mikolov, Ilya Sutskever, Kai Chen, Greg~S Corrado, and Jeff Dean.
  2013{\natexlab{b}}.
\newblock
  \href{http://papers.nips.cc/paper/5021-distributed-representations-of-words-and-phrases-and-their-compositionality.pdf}{Distributed
  representations of words and phrases and their compositionality}.
\newblock In {\em Proceedings of NIPS 2013\/}.
\newblock
  \href{http://papers.nips.cc/paper/5021-distributed-representations-of-words-and-phrases-and-their-compositionality.pdf}{http://papers.nips.cc/paper/5021-distributed-representations-of-words-and-phrases-and-their-compositionality.pdf}.

\bibitem[{Miller(1995)}]{Miller1995}
George~A. Miller. 1995.
\newblock \href{https://doi.org/10.1145/219717.219748}{{WordNet}: A lexical
  database for english}.
\newblock {\em Commun. ACM\/} 38(11):39--41.
\newblock
  \href{https://doi.org/10.1145/219717.219748}{https://doi.org/10.1145/219717.219748}.

\bibitem[{Moschitti(2006)}]{Moschitti:2006}
Alessandro Moschitti. 2006.
\newblock \href{http://dx.doi.org/10.1007/11871842\_32}{Efficient convolution
  kernels for dependency and constituent syntactic trees}.
\newblock In {\em Proceedings of ECML'06\/}.
\newblock
  \href{http://dx.doi.org/10.1007/11871842\_32}{http://dx.doi.org/10.1007/11871842\_32}.

\bibitem[{Mu et~al.(2017)Mu, Bhat, and Viswanath}]{Mu2017}
Jiaqi Mu, Suma Bhat, and Pramod Viswanath. 2017.
\newblock \href{http://arxiv.org/abs/1704.05358}{Representing sentences as
  low-rank subspaces}.
\newblock {\em CoRR\/} abs/1704.05358.
\newblock
  \href{http://arxiv.org/abs/1704.05358}{http://arxiv.org/abs/1704.05358}.

\bibitem[{Nagoudi et~al.(2017)Nagoudi, Ferrero, and
  Schwab}]{nagoudi-ferrero-schwab:2017:SemEval}
El~Moatez~Billah Nagoudi, J\'{e}r\'{e}my Ferrero, and Didier Schwab. 2017.
\newblock \href{http://www.aclweb.org/anthology/S17-2017}{{LIM-LIG at
  SemEval-2017 Task1}: Enhancing the semantic similarity for arabic sentences
  with vectors weighting}.
\newblock In {\em Proceedings of SemEval-2017\/}.
\newblock
  \href{http://www.aclweb.org/anthology/S17-2017}{http://www.aclweb.org/anthology/S17-2017}.

\bibitem[{Navigli and Ponzetto(2010)}]{Navigli:2010:BBV:1858681.1858704}
Roberto Navigli and Simone~Paolo Ponzetto. 2010.
\newblock \href{http://aclweb.org/anthology/P/P10/P10-1023.pdf}{{BabelNet}:
  Building a very large multilingual semantic network}.
\newblock In {\em Proceedings of ACL 2010\/}.
\newblock
  \href{http://aclweb.org/anthology/P/P10/P10-1023.pdf}{http://aclweb.org/anthology/P/P10/P10-1023.pdf}.

\bibitem[{Pagliardini et~al.(2017)Pagliardini, Gupta, and Jaggi}]{pgj2017unsup}
Matteo Pagliardini, Prakhar Gupta, and Martin Jaggi. 2017.
\newblock \href{https://arxiv.org/pdf/1703.02507.pdf}{{Unsupervised Learning of
  Sentence Embeddings using Compositional n-Gram Features}}.
\newblock {\em arXiv\/}
  \href{https://arxiv.org/pdf/1703.02507.pdf}{https://arxiv.org/pdf/1703.02507.pdf}.

\bibitem[{Parker et~al.(2011)Parker, Graff, Kong, Chen, and
  Maeda}]{parker-EtAl:2011}
Robert Parker, David Graff, Junbo Kong, Ke~Chen, and Kazuaki Maeda. 2011.
\newblock {\em Gigaword Fifth Edition LDC2011T07\/}.
\newblock Linguistic Data Consortium.
\newblock
  \href{https://catalog.ldc.upenn.edu/ldc2011t07}{https://catalog.ldc.upenn.edu/ldc2011t07}.

\bibitem[{Pennington et~al.(2014)Pennington, Socher, and
  Manning}]{pennington2014glove}
Jeffrey Pennington, Richard Socher, and Christopher~D. Manning. 2014.
\newblock \href{http://www.aclweb.org/anthology/D14-1162}{{GloVe}: {G}lobal
  {V}ectors for {W}ord {R}epresentation}.
\newblock In {\em Proceedings of EMNLP 2014\/}.
\newblock
  \href{http://www.aclweb.org/anthology/D14-1162}{http://www.aclweb.org/anthology/D14-1162}.

\bibitem[{Pham et~al.(2015)Pham, Kruszewski, Lazaridou, and Baroni}]{pham2015}
Nghia~The Pham, Germ\'{a}n Kruszewski, Angeliki Lazaridou, and Marco Baroni.
  2015.
\newblock \href{http://www.aclweb.org/anthology/P15-1094}{Jointly optimizing
  word representations for lexical and sentential tasks with the c-phrase
  model}.
\newblock In {\em Proceedings of ACL/IJCNLP\/}.
\newblock
  \href{http://www.aclweb.org/anthology/P15-1094}{http://www.aclweb.org/anthology/P15-1094}.

\bibitem[{Reimers et~al.(2016)Reimers, Beyer, and
  Gurevych}]{reimers-beyer-gurevych:2016:COLING}
Nils Reimers, Philip Beyer, and Iryna Gurevych. 2016.
\newblock \href{http://aclweb.org/anthology/C16-1009}{Task-oriented intrinsic
  evaluation of semantic textual similarity}.
\newblock In {\em Proceedings of COLING 2016\/}.
\newblock
  \href{http://aclweb.org/anthology/C16-1009}{http://aclweb.org/anthology/C16-1009}.

\bibitem[{Rychalska et~al.(2016)Rychalska, Pakulska, Chodorowska, Walczak, and
  Andruszkiewicz}]{Rychalska2016}
Barbara Rychalska, Katarzyna Pakulska, Krystyna Chodorowska, Wojciech Walczak,
  and Piotr Andruszkiewicz. 2016.
\newblock \href{http://www.aclweb.org/anthology/S16-1091}{{S}amsung {P}oland
  {NLP} {T}eam at {S}em{E}val-2016 {T}ask 1: Necessity for diversity; combining
  recursive autoencoders, wordnet and ensemble methods to measure semantic
  similarity.}
\newblock In {\em Proceedings of SemEval-2016\/}.
\newblock
  \href{http://www.aclweb.org/anthology/S16-1091}{http://www.aclweb.org/anthology/S16-1091}.

\bibitem[{Salle et~al.(2016{\natexlab{a}})Salle, Idiart, and
  Villavicencio}]{Salle2016arxiv}
Alexandre Salle, Marco Idiart, and Aline Villavicencio. 2016{\natexlab{a}}.
\newblock \href{http://arxiv.org/abs/1606.01283}{Enhancing the lexvec
  distributed word representation model using positional contexts and external
  memory}.
\newblock {\em CoRR\/} abs/1606.01283.
\newblock
  \href{http://arxiv.org/abs/1606.01283}{http://arxiv.org/abs/1606.01283}.

\bibitem[{Salle et~al.(2016{\natexlab{b}})Salle, Idiart, and
  Villavicencio}]{Salle2016acl}
Alexandre Salle, Marco Idiart, and Aline Villavicencio. 2016{\natexlab{b}}.
\newblock \href{http://aclweb.org/anthology/P16-2068}{Matrix factorization
  using window sampling and negative sampling for improved word
  representations}.
\newblock In {\em Proceedings of ACL\/}.
\newblock
  \href{http://aclweb.org/anthology/P16-2068}{http://aclweb.org/anthology/P16-2068}.

\bibitem[{Serasset(2015)}]{Serasset2015}
Gilles Serasset. 2015.
\newblock \href{https://doi.org/10.3233/SW-140147}{{DBnary}: Wiktionary as a
  lemon-based multilingual lexical resource in {RDF}}.
\newblock {\em Semantic Web Journal (special issue on Multilingual Linked Open
  Data)\/} 6:355–--361.
\newblock
  \href{https://doi.org/10.3233/SW-140147}{https://doi.org/10.3233/SW-140147}.

\bibitem[{Shao(2017)}]{shao:2017:SemEval}
Yang Shao. 2017.
\newblock \href{http://www.aclweb.org/anthology/S17-2016}{{HCTI at SemEval-2017
  Task 1}: Use convolutional neural network to evaluate semantic textual
  similarity}.
\newblock In {\em Proceedings of SemEval-2017\/}.
\newblock
  \href{http://www.aclweb.org/anthology/S17-2016}{http://www.aclweb.org/anthology/S17-2016}.

\bibitem[{Shen et~al.(2014)Shen, He, Gao, Deng, and Mesnil}]{YelongShen2014}
Yelong Shen, Xiaodong He, Jianfeng Gao, Li~Deng, and Gregoire Mesnil. 2014.
\newblock
  \href{https://www.microsoft.com/en-us/research/publication/a-latent-semantic-model-with-convolutional-pooling-structure-for-information-retrieval/}{A
  latent semantic model with convolutional-pooling structure for information
  retrieval}.
\newblock In {\em Proceedings of CIKM '14\/}.
\newblock
  \href{https://www.microsoft.com/en-us/research/publication/a-latent-semantic-model-with-convolutional-pooling-structure-for-information-retrieval/}{https://www.microsoft.com/en-us/research/publication/a-latent-semantic-model-with-convolutional-pooling-structure-for-information-retrieval/}.

\bibitem[{Snover et~al.(2006)Snover, Dorr, Schwartz, Micciulla, and
  Makhoul}]{snover2006}
Matthew Snover, Bonnie Dorr, Richard Schwartz, Linnea Micciulla, and John
  Makhoul. 2006.
\newblock \href{http://mt-archive.info/AMTA-2006-Snover.pdf}{A study of
  translation edit rate with targeted human annotation}.
\newblock In {\em Proceedings of AMTA 2006\/}.
\newblock
  \href{http://mt-archive.info/AMTA-2006-Snover.pdf}{http://mt-archive.info/AMTA-2006-Snover.pdf}.

\bibitem[{\'{S}piewak et~al.(2017)\'{S}piewak, Sobecki, and
  Kara\'{s}}]{spiewak-sobecki-karas:2017:SemEval}
Martyna \'{S}piewak, Piotr Sobecki, and Daniel Kara\'{s}. 2017.
\newblock \href{http://www.aclweb.org/anthology/S17-2018}{{OPI-JSA at
  SemEval-2017 Task 1}: Application of ensemble learning for computing semantic
  textual similarity}.
\newblock In {\em Proceedings of SemEval-2017\/}.
\newblock
  \href{http://www.aclweb.org/anthology/S17-2018}{http://www.aclweb.org/anthology/S17-2018}.

\bibitem[{Sultan et~al.(2015)Sultan, Bethard, and Sumner}]{Sultan2015}
Md~Arafat Sultan, Steven Bethard, and Tamara Sumner. 2015.
\newblock \href{http://aclweb.org/anthology/S/S15/S15-2027.pdf}{{DLS@CU}:
  Sentence similarity from word alignment and semantic vector composition}.
\newblock In {\em Proceedings of SemEval 2015\/}.
\newblock
  \href{http://aclweb.org/anthology/S/S15/S15-2027.pdf}{http://aclweb.org/anthology/S/S15/S15-2027.pdf}.

\bibitem[{Tian et~al.(2017)Tian, Zhou, Lan, and Wu}]{tian-EtAl:2017:SemEval}
Junfeng Tian, Zhiheng Zhou, Man Lan, and Yuanbin Wu. 2017.
\newblock \href{http://www.aclweb.org/anthology/S17-2028}{{ECNU at SemEval-2017
  Task 1}: Leverage kernel-based traditional nlp features and neural networks
  to build a universal model for multilingual and cross-lingual semantic
  textual similarity}.
\newblock In {\em Proceedings of SemEval-2017\/}.
\newblock
  \href{http://www.aclweb.org/anthology/S17-2028}{http://www.aclweb.org/anthology/S17-2028}.

\bibitem[{\v{S}ari\'{c} et~al.(2012{\natexlab{a}})\v{S}ari\'{c}, Glava\v{s},
  Karan, \v{S}najder, and
  Dalbelo~Ba\v{s}i\'{c}}]{vsaric-EtAl:2012:STARSEM-SEMEVAL}
Frane \v{S}ari\'{c}, Goran Glava\v{s}, Mladen Karan, Jan \v{S}najder, and
  Bojana Dalbelo~Ba\v{s}i\'{c}. 2012{\natexlab{a}}.
\newblock \href{http://www.aclweb.org/anthology/S12-1060}{Takelab: Systems for
  measuring semantic text similarity}.
\newblock In {\em Proceedings of {*SEM 2012}/{SemEval 2012}\/}.
\newblock
  \href{http://www.aclweb.org/anthology/S12-1060}{http://www.aclweb.org/anthology/S12-1060}.

\bibitem[{\v{S}ari\'{c} et~al.(2012{\natexlab{b}})\v{S}ari\'{c}, Glava\v{s},
  Karan, \v{S}najder, and Dalbelo~Ba\v{s}i\'{c}}]{saric2012takelab}
Frane \v{S}ari\'{c}, Goran Glava\v{s}, Mladen Karan, Jan \v{S}najder, and
  Bojana Dalbelo~Ba\v{s}i\'{c}. 2012{\natexlab{b}}.
\newblock \href{http://www.aclweb.org/anthology/S12-1060}{{TakeLab: Systems for
  measuring semantic text similarity}}.
\newblock In {\em Proceedings of SemEval 2012\/}.
\newblock
  \href{http://www.aclweb.org/anthology/S12-1060}{http://www.aclweb.org/anthology/S12-1060}.

\bibitem[{Wieting et~al.(2015)Wieting, Bansal, Gimpel, and
  Livescu}]{Wieting2015tacl}
John Wieting, Mohit Bansal, Kevin Gimpel, and Karen Livescu. 2015.
\newblock \href{http://aclweb.org/anthology/Q/Q15/Q15-1025.pdf}{From paraphrase
  database to compositional paraphrase model and back}.
\newblock {\em Transactions of the ACL (TACL)\/} 3:345--358.
\newblock
  \href{http://aclweb.org/anthology/Q/Q15/Q15-1025.pdf}{http://aclweb.org/anthology/Q/Q15/Q15-1025.pdf}.

\bibitem[{Wieting et~al.(2016{\natexlab{a}})Wieting, Bansal, Gimpel, and
  Livescu}]{wieting-EtAl:2016:EMNLP2016}
John Wieting, Mohit Bansal, Kevin Gimpel, and Karen Livescu.
  2016{\natexlab{a}}.
\newblock \href{https://aclweb.org/anthology/D16-1157}{Charagram: Embedding
  words and sentences via character n-grams}.
\newblock In {\em Proceedings of EMNLP\/}.
\newblock
  \href{https://aclweb.org/anthology/D16-1157}{https://aclweb.org/anthology/D16-1157}.

\bibitem[{Wieting et~al.(2016{\natexlab{b}})Wieting, Bansal, Gimpel, and
  Livescu}]{Wieting2016}
John Wieting, Mohit Bansal, Kevin Gimpel, and Karen Livescu.
  2016{\natexlab{b}}.
\newblock \href{http://arxiv.org/abs/1511.08198}{Towards universal paraphrastic
  sentence embeddings}.
\newblock In {\em Proceedings of ICLR 2016\/}.
\newblock
  \href{http://arxiv.org/abs/1511.08198}{http://arxiv.org/abs/1511.08198}.

\bibitem[{Wieting and Gimpel(2017)}]{Wieting2017}
John Wieting and Kevin Gimpel. 2017.
\newblock \href{http://arxiv.org/abs/1705.00364}{Revisiting recurrent networks
  for paraphrastic sentence embeddings}.
\newblock {\em CoRR\/} abs/1705.00364.
\newblock
  \href{http://arxiv.org/abs/1705.00364}{http://arxiv.org/abs/1705.00364}.

\bibitem[{Wu et~al.(2017)Wu, Huang, Jian, Guo, and Su}]{wu-EtAl:2017:SemEval1}
Hao Wu, Heyan Huang, Ping Jian, Yuhang Guo, and Chao Su. 2017.
\newblock \href{http://www.aclweb.org/anthology/S17-2007}{{BIT at SemEval-2017
  Task 1}: Using semantic information space to evaluate semantic textual
  similarity}.
\newblock In {\em Proceedings of SemEval-2017\/}.
\newblock
  \href{http://www.aclweb.org/anthology/S17-2007}{http://www.aclweb.org/anthology/S17-2007}.

\bibitem[{Xu et~al.(2015)Xu, Callison-Burch, and Dolan}]{xu2015semeval}
Wei Xu, Chris Callison-Burch, and William~B. Dolan. 2015.
\newblock \href{http://www.aclweb.org/anthology/S15-2001}{{SemEval-2015 Task
  1}: Paraphrase and semantic similarity in {Twitter} ({PIT})}.
\newblock In {\em Proceedings of SemEval 2015\/}.
\newblock
  \href{http://www.aclweb.org/anthology/S15-2001}{http://www.aclweb.org/anthology/S15-2001}.

\bibitem[{Young et~al.(2014)Young, Lai, Hodosh, and Hockenmaier}]{young2014}
Peter Young, Alice Lai, Micah Hodosh, and Julia Hockenmaier. 2014.
\newblock \href{http://aclweb.org/anthology/Q14-1006}{From image descriptions
  to visual denotations: New similarity metrics for semantic inference over
  event descriptions}.
\newblock {\em TACL\/} 2:67--78.
\newblock
  \href{http://aclweb.org/anthology/Q14-1006}{http://aclweb.org/anthology/Q14-1006}.

\bibitem[{Zhuang and Chang(2017)}]{zhuang-chang:2017:SemEval}
WenLi Zhuang and Ernie Chang. 2017.
\newblock \href{http://www.aclweb.org/anthology/S17-2023}{{Neobility at
  SemEval-2017 Task 1}: An attention-based sentence similarity model}.
\newblock In {\em Proceedings SemEval-2017\/}.
\newblock
  \href{http://www.aclweb.org/anthology/S17-2023}{http://www.aclweb.org/anthology/S17-2023}.

\end{thebibliography}
\bibliographystyle{acl_natbib}

\end{document}